%% file: main.tex
\documentclass[10pt,twocolumn,letterpaper]{article}

\usepackage[pagenumbers]{cvpr} % To force page numbers, e.g. for an arXiv version

\input{preamble}

\usepackage{multirow}
\usepackage{comment}

\newcommand\figref[1]{{Fig.~\color{red}\ref{#1}}}
\newcommand\tabref[1]{{Table~\color{red}\ref{#1}}}
\newcommand\secref[1]{{Sec.~\color{red}\ref{#1}}}

\definecolor{cvprblue}{rgb}{0.21,0.49,0.74}
\usepackage[pagebackref,breaklinks,colorlinks,citecolor=cvprblue]{hyperref}

\def\paperID{12431} % *** Enter the Paper ID here
\def\confName{CVPR}
\def\confYear{2024}

\title{NoiseCollage: A Layout-Aware Text-to-Image Diffusion Model\\ Based on Noise Cropping and Merging}

\author{Takahiro Shirakawa, Seiichi Uchida\\
Kyushu University, Japan\\
{\tt\small takahiro.shirakawa@human.ait.kyushu-u.ac.jp, uchida@ait.kyushu-u.ac.jp}
}

\begin{document}
\maketitle

\begin{abstract}
Layout-aware text-to-image generation is a task to generate multi-object images that reflect layout conditions in addition to text conditions.
The current layout-aware text-to-image diffusion models still have several issues, including mismatches between the text and layout conditions and quality degradation of generated images.
This paper proposes a novel layout-aware text-to-image diffusion model called NoiseCollage to tackle these issues.
During the denoising process, NoiseCollage independently estimates noises for individual objects and then crops and merges them into a single noise.
This operation helps avoid condition mismatches; in other words, it can put the right objects in the right places. Qualitative and quantitative evaluations show that NoiseCollage outperforms several state-of-the-art models. These successful results indicate that the crop-and-merge operation of noises is a reasonable strategy to control image generation. 
We also show that NoiseCollage can be integrated with ControlNet to use edges, sketches, and pose skeletons as additional conditions. 
Experimental results show that this integration boosts the layout accuracy of ControlNet.
The code is available at \url{https://github.com/univ-esuty/noisecollage}.
\end{abstract}

\section{Introduction} \label{sec:intro}
% 全体（グローバル）背景
Diffusion models, such as StableDiffusion (SD)~\cite{sd}, have rapidly improved text-to-image generation. In general, diffusion models generate images through a denoising process, an iterative process to remove noise from an initial Gaussian noise image. 
A UNet estimates the noise with a text condition.
%and removed from the input image, as shown in \figref{fig:overview}~(a). 
After the denoising iterations, the model gives a noise-free (clean) image that reflects the text condition.
\par

\begin{figure}[t]
\centering
\includegraphics[width=\linewidth]{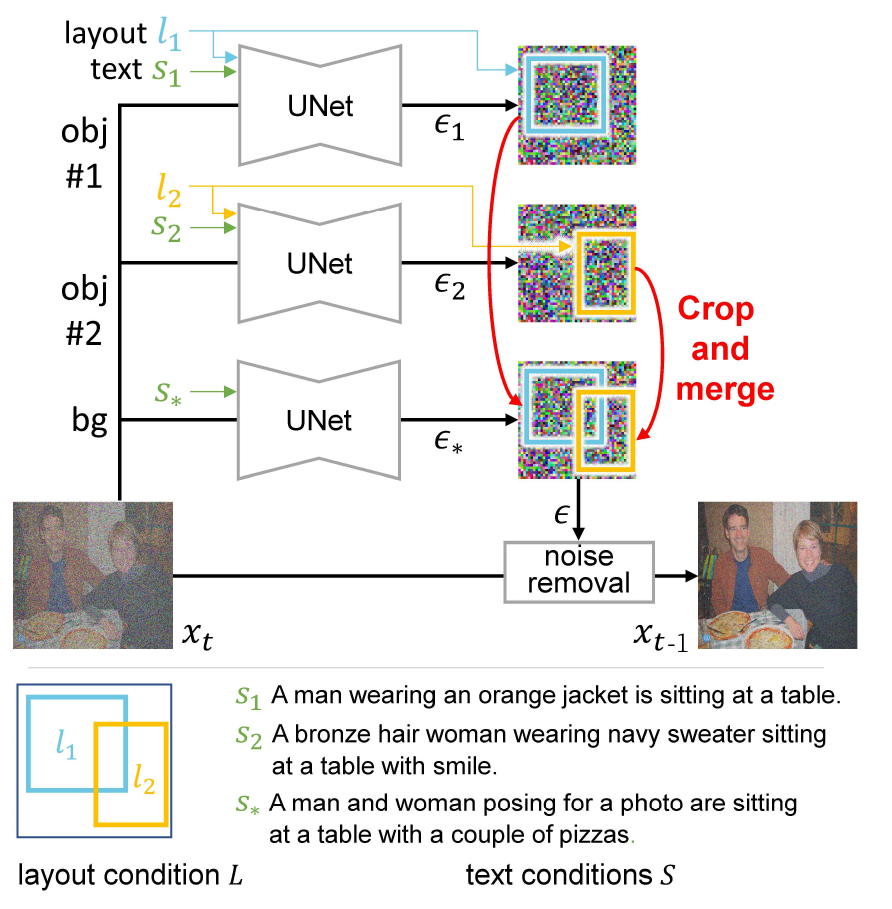}
\caption{Denoising processes of NoiseCollage. (Although illustrated as a process in the image space, the actual denoising process is performed in a latent space like~\cite{sd} for computational efficiency.)}
\label{fig:overview}
\end{figure}

Text-to-image diffusion models have recently been extended to generate multiple objects with layout awareness. Namely, these models can generate images with multiple objects while controlling their spatial locations. There are two approaches for the extension: attention manipulation~\cite{ediff,zestguide,boxdiffusion,layout-cyber1,layout-cyber2,Kim2023DenseTG,Xiao2023RBRA,Phung2023GroundedTS} and iterative image editing~\cite{collage-diffusion,continuous-layout,paint-by-example,blend-dm,blend-ldm,Zhang2023paste}. The former manipulates the cross attention layers in the UNet for letting a certain region only focus on a certain object.
The latter generates an initial image and then puts another object in the initial image. More objects can be arranged by repeating this editing process.
\par
The current layout-aware text-to-image diffusion models still have the following limitations. Specifically, the first approach, attention manipulation, often shows mismatches between the text and layout conditions. The second approach, iterative editing, shows the image quality degradation as it iterates to show more objects.
\par

% 提案手法
This paper proposes a novel layout-aware text-to-image diffusion model called {\em NoiseCollage}. 
\figref{fig:overview} shows an overview of the denoising process of NoiseCollage. When generating images with $N$ objects, NoiseCollage takes $N+1$ text conditions (i.e., prompts) $\{s_1,\ldots, s_N, s_\ast\}$ and $N$ layout conditions $\{l_1, \ldots, l_N\}$ for image generation. Namely, a pair of text and layout conditions $(s_n, l_n)$ are given for each object $n$. As shown in \figref{fig:overview}, each layout condition is specified by a bounding box unless otherwise mentioned. The remaining text condition $s_{\ast}$ roughly describes the whole image. \par

% 提案手法の一番の売りポイント(マルチテキスト・適当なレイアウト作成・同時生成・トレーニングフリー)
The technical highlight of NoiseCollage is that $N+1$ noises $\{\epsilon_1, \ldots, \epsilon_N, \epsilon_\ast\}$ for $N$ objects and the whole image are estimated independently and then assembled like image collage. More specifically, the region $l_n$ for the $n$-th object is cropped from $\epsilon_n$, and then the $N$ cropped noises are merged with the noise for the whole image, $\epsilon_\ast$. This operation is novel and very different from the existing text-to-image diffusion models; our assemblage operation {\em directly} creates a noise from $N+1$ noises to have an expected output image. In other words, our trials on NoiseCollage indicate that {\em multi-object images can be generated accurately by the crop-and-merge operation of noises}. \par
For accurate and flexible image generation, we introduce three gimmicks in NoiseCollage. The first gimmick is {\em masked cross attention}. This gimmick aims to estimate a noise $\epsilon_n$ that accurately reflects
the text condition $s_n$ around the region $l_n$. The second gimmick is to make the crop-and-merge operation to be soft. More specifically, we use a weighted merging operation so that the cropped noises do not completely overwrite the global information of $\epsilon_\ast$. The weighted merging operation also allows (even large) overlaps between the regions $\{l_n\}$. The third gimmick is an integration of ControlNet\cite{controlnet} to allow more flexible conditions. ControlNet employs various conditions, such as pose skeleton and edge images, for guiding image generation; therefore, the integration with ControlNet allows NoiseCollage to use these guiding conditions.
\par

Like other popular layout-aware image generation methods~\cite{ediff,collage-diffusion}, NoiseCollage is {\em training-free} and thus can employ various diffusion models pre-trained to generate images from a text condition. We mainly used a pretrained SD for photo-realistic images in the later experiments. However, as noted above, we also used an SD model for anime images and ControlNet. If we have better diffusion models in the near future, we can employ them for NoiseCollage without any modification.\par
%
% NoiseCollage also has three practical advantages, each contributing to generating higher-quality images. First, since it independently estimates noises $\{\epsilon_1,\ldots,\epsilon_N\}$, the text and layout conditions are directly and accurately reflected on the corresponding objects. This means it is possible to minimize the risk of confusing the correspondence between the objects and the conditions, whereas the existing models based on attention manipulation often suffer from it. Second, NoiseCollage generates an image containing all $N$ objects by the single iterative process and, therefore, 
% can avoid the degradation of objects generated in earlier iterations, which occurs in the iterative editing models. Third,

%
% 実験について
We conduct various qualitative and quantitative evaluation experiments to confirm that our NoiseCollage outperforms the state-of-the-art layout-aware image generation models. We first observe that NoiseCollage generates multi-object images that are high-quality and accurate to the input conditions. We then quantitatively evaluate how accurately the given conditions are reflected in the corresponding objects. For this evaluation, we introduce multimodal feature representation by CLIP\cite{clip}; if a model shows high CLIP-feature similarity between text conditions and generated images, the model will have high accuracy to the input conditions. \par
The main contributions of this paper are summarized as follows.
\begin{itemize}
\item We propose NoiseCollage, a novel layout-aware text-to-image diffusion model. It can generate multi-object images that accurately reflect text and layout conditions.
\item NoiseCollage is the first method that performs a crop-and-merge operation of noises estimated for individual objects in its denoising process. Its accurate and high-quality generated images without artifacts indicate that noise is a good medium for direct layout control.
\item Experimental results show that NoiseCollage outperforms the state-of-the-art methods by avoiding condition mismatches.
\item The Training-free nature of NoiseCollage allows direct integration with ControlNet and realizes finer output controls by edge images, sketches, and body skeletons.
\end{itemize}

%==================================================================
\section{Related Work} \label{sec:related}
%==================================================================
%------------------------------------------------------------------
\subsection{Text-to-Image Diffusion Models}
% diffusion model全般の話
Many image generation methods based on diffusion models have been proposed so far~\cite{ddpm-org,ddpm,ddim,improved-ddpm,dmbg}. For generating high-resolution images without a drastic increase in computational costs, they often employ the technique of Latent Diffusion Model (LDM)~\cite{sd}, where the denoising process with UNet runs in a low-dimensional latent space. Various conditions are also introduced to control the generated images.\par
%
% text-to-image diffusion model全般の話
Text-to-image diffusion models~\cite{sd,sdxl,dalle2,imagen,ediff,make-a-scene} can generate high-quality and diverse images with a text condition. Among them, StableDiffusion (SD)~\cite{sd} is one of the most popular models. Those models have been extended to realize other image processing with tasks, such as image editing~\cite{Couairon2022DiffEditDS,Hertz2022PrompttoPromptIE,Brooks2022InstructPix2PixLT}, image inpainting ~\cite{Lugmayr2022RePaintIU,Zhang2023TowardsCI,Xie2022SmartBrushTA}, and image-to-image translation~\cite{palette,sdedit,Parmar2023ZeroshotIT}.

%------------------------------------------------------------------
\subsection{Layout-Aware Diffusion Models}
Layout-aware text-to-image generation is a task to generate multi-object images that reflect a layout condition in addition to a text condition.
Several fine-tuning techniques~\cite{controlnet,spatext,Cheng2023LayoutDiffuseAF,Xue2023FreestyleLS,Jia2023SSMGSM} have been proposed to incorporate the layout condition into the pre-trained diffusion model. For example, MultiDiffusion employs an extra optimization step to mix the denoised images into one. ControlNet~\cite{controlnet} combines SD and a trainable encoder of various conditions for fine layout control, such as pose skeletons for human pose control.\par
%
% non-iterative manner
We can find {\em training-free} methods that use the pre-trained models without fine-tuning steps for layout conditions. They are classified into {\em attention manipulation} and {\em iterative editing}.
Attention manipulation methods~\cite{ediff,zestguide,boxdiffusion,layout-cyber1,layout-cyber2,Kim2023DenseTG,Xiao2023RBRA,Phung2023GroundedTS} 
control the object layout by manipulating a cross-attention layer, which is an important module in UNet to correlate text conditions and regions in the generated images. Paint-with-words\cite{ediff} is the most popular state-of-the-art method that uses attention manipulation. It can generate images from a text condition and an object segmentation mask. A word (such as ``rabbit'') in the text condition is given to each segment, and this word-segment correspondence is then used to modify the cross-attention. However, as we will see later, controlling the correspondence between multiple objects and their regions within a cross-attention layer is tough and often suffers from wrong correspondences, i.e., condition mismatches.  
\par 
%
% iterative manner
Iterative editing\cite{collage-diffusion,continuous-layout,paint-by-example,blend-dm,blend-ldm,Zhang2023paste} is a more intuitive way to deal with multiple objects and their layout. Given a pre-generated initial image, one object is placed at its position, and then the next object is placed. Repeating this step $N$ times gives us an image with $N$ objects.
Collage Diffusion~\cite{collage-diffusion} is a popular iterative editing method. Although it introduces an extra diffusion and denoising step like 
SDEdit\cite{sdedit} to harmonize the newly added object in the resulting image, it still suffers from image quality degradation, which becomes more serious according to many iterations.\par
\tabref{tab:func-comparison} summarizes functionality comparisons between popular layout-aware text-to-image methods and our NoiseCollage. ``Multi-prompts'' is the function to accept different text conditions for individual objects. ``Region overlap'' is the function to allow overlapping layout conditions (by, for example, bounding boxes) for objects. As indicated by this table, our NoiseCollage has several promising properties.\par

\begin{table}[t]
\centering
\caption{Comparison of popular state-of-the-art layout-aware text-to-image diffusion models.}
\scalebox{.85}[.85]{
\begin{tabular}{l|ccc|c}
\hline
              & Control-        & Paint-with               & Collage                  & Ours            \\
                            & Net\cite{controlnet}         & words\cite{ediff}               & Diffusion\cite{collage-diffusion}                  &           \\
\hline
training-free & $\times$        & $\surd$             & $\surd$             & $\surd$         \\
non-iterative & $\surd$         & $\surd$             & $\times$            & $\surd$         \\
multi-prompts & $\times$        & $\times$            & $\surd$             & $\surd$         \\
region overlap& $\surd$         & $\times$            & $\surd$            & $\surd$         \\
\hline                
\end{tabular}
}
\label{tab:func-comparison}
\end{table}
%
%----------------------------------------------------------------------------
\subsection{Noise Manipulation}
The most popular manipulation of the estimated noise in diffusion models is classifier-free guidance~\cite{cfg}. It uses the difference between a pair of noises estimated with and without a class condition. This difference reflects the class-specific characteristics and thus is useful to emphasize them in the generated images.\par
To the authors' knowledge, no existing model manipulates the estimated noises in a direct manner, such as our crop-and-merge operation. As we will see in this paper, noises are a good medium for allowing simple and intuitive manipulations to control the object layout without introducing any artifacts.
% 8:30AM 15th Nov.
%==================================================================
\section{NoiseCollage} \label{sec:proposed}
%==================================================================
%------------------------------------------------------------------
\subsection{Overview} \label{sec:overview}
NoiseCollage generates an image with $N$ objects from the following conditions, $L$, $S$, and $s_{\ast}$: 
\begin{itemize}
    \item $L=\{l_1,\ldots,l_N\}$ is the $N$ layout conditions to control the layout of individual objects. Each layout condition $l_n$ is represented as a region specified by a bounding box or a polygon. Note that regions can be overlapped; thus, there is no need to be nervous about setting layout conditions.
    \item $S=\{s_1,\ldots,s_N\}$ is the set of $N$ text conditions to describe the visual information of the objects. Each condition is given as a word sequence; for example, ``A man wearing an orange jacket is sitting at a table.''
    \item $s_{\ast}$ is a global text condition to describe the whole image of the objects. Although we call it ``global,''  $s_{\ast}$ need {\em not} describe everything. The global text $s_{\ast}$ may outline the whole image or include descriptions of several objects.
\end{itemize}
\par
NoiseCollage uses the denoising process of standard diffusion models. NoiseCollage starts from $t=T$ with a Gaussian noise image $x_T$. Then, from $t=T$ to $1$, it uses a pre-trained UNet to estimate the noise $\epsilon$ at each $t$ from a noisy image $x_t$ and then removes the noise $\epsilon$ from $x_t$ to have a less-noisy image $x_{t-1}$. The denoising process finally provides an image $x_0$, which satisfies the given conditions.\par
% 22:00 Nov.13
%
%\figref{fig:proposed/abstract} illustrates
%
The main difference between NoiseCollage and the standard diffusion models is that it derives the noise $\epsilon$ at $t$ by a crop-and-merge operation (i.e., collage) of $N+1$ noises, $\{\epsilon_1,\ldots,\epsilon_N, \epsilon_\ast\}$\footnote{Precisely speaking, the noise $\epsilon$ should be denoted as $\epsilon(x_t\ |\ t, L, S, s_\ast)$, because it is estimated from $x_t$ at timestep $t$  under the conditions $L$, $S$, and $s_\ast$. Similarly,  $\epsilon_n$ and $\epsilon_\ast$ are denoted as $\epsilon_n(x_t\ |\ t, l_n, s_n)$ and $\epsilon_\ast(x_t\ |\ t, s_\ast)$, respectively. In this paper, we use the simpler notation $\epsilon$, $\epsilon_n$, and $\epsilon_\ast$, unless there is confusion.}, as shown in \figref{fig:overview}.
Roughly speaking, the noise $\epsilon$ is given by cropping the region specified by $l_n$ from $\epsilon_n$ for each $n$ and then merging the $N$ cropped regions with $\epsilon_\ast$. In the following, Section~\ref{sec:cam} details the crop-and-merge operation. Then, Section~\ref{sec:mca} details the masked cross-attention mechanism, which is necessary to make the crop-and-merge operation work as expected.
\par
% 0:00AM Nov 14th SU
%------------------------------------------------------------------
\subsection{Crop-and-Merge Operation of Noises} \label{sec:cam}
A naive crop-and-merge operation for creating $\epsilon$ is to use $\epsilon_n$ for the $n$-th object region $l_n$ and $\epsilon_\ast$ for the region for the non-object region.
However, this naive operation has two issues. First, $\epsilon_\ast$ should not be excluded from the object region $l_n$. For example, when generating a ring-shaped object in the box $l_n$, $\epsilon_\ast$ is necessary for the non-ring area within $l_n$. Second, the naive operation does not consider the overlapping regions among $\{l_n\}$.\par
\begin{figure}[t]
\centering
\includegraphics[width=\linewidth]{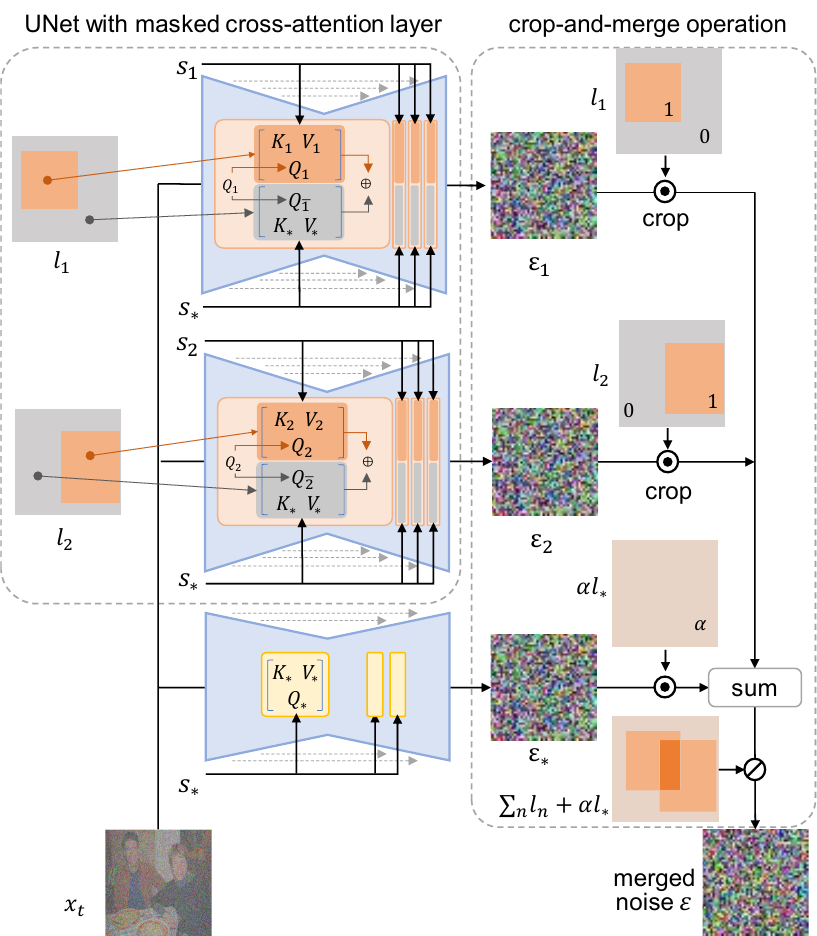}
\caption{Overview of the noise estimation process in our NoiseCollage.}
\label{fig:abstract}
\end{figure}

We, therefore, use the following crop-and-merge operation, illustrated in the right side of \figref{fig:abstract}:
\begin{equation}
\epsilon = ({\textstyle\sum_n} l_n\epsilon_n + \alpha l_\ast\epsilon_\ast) / (\textstyle{\sum_n} l_n  +\alpha l_\ast). \label{eq:c-a-m}
\end{equation}
Here, $l_n$ is treated as a binary mask image whose pixel value is 1 for the region specified by $l_n$, and $l_\ast$ is an image whose all pixels are 1. In Eq.~\ref{eq:c-a-m}, addition, multiplication, and division are pixel-wise. The hyper-parameter $\alpha$ is the weight to control the strength of $\epsilon_\ast$ within the object regions and set at 0.1 by a preliminary experiment. 

%------------------------------------------------------------------
\subsection{Masked Cross-Attention} \label{sec:mca}
The cross-attention layer in the UNet of standard text-to-image diffusion models is an important module to correlate texts and image regions. Specifically, it calculates $\tilde{Q}=\mathrm{softmax}({QK^{T}}/{\sqrt{d}}){V}$, where the query $Q$ is a matrix with $N$ $d$-dimensional image features of $x_t$, whereas the key $K$ and the value $V$ are the same matrix with $M$ $d$-dimensional text 
features from the text conditions $(S, s_\ast)$. Through this layer, $N$ image features $Q$ are converted into $N$ image features (denoted as $\tilde{Q}$) that reflect text conditions.\par
We propose a ``masked'' cross-attention layer, which is a simple extension of the above cross-attention, as shown in the left side of \figref{fig:abstract}.
In the process of estimating $\epsilon_n$ in NoiseCollage, the visual information of the $n$-th object by $s_n$ should be localized around the region $l_n$ in $\epsilon_n$; this is because only the region $l_n$ is cropped and merged into $\epsilon$. 
For this localization, we split the cross-attention operation into two sub-operations: one is 
a sub-operation to correlate the region $l_n$ with $s_n$ and the other to correlate the remaining region with $s_\ast$. Specifically, we first derive two ``masked'' matrices $Q_n$ and $Q_{\overline{n}}$ from $Q$, where the matrix $Q_n$ has the value of $Q$ at the columns corresponding to $l_n$ and zero at the other columns and $Q_{\overline{n}}=Q\ominus Q_n$.  We also derive $V_n$ and $K_n$ from $s_n$ and $V_{\ast}$ from $s_\ast$.  We then get the cross-attention results for the $n$-th object as $\tilde{Q}_n=\mathrm{softmax}({Q_nK_n^{T}}/{\sqrt{d}}){V_n}$ and for the other region as $\tilde{Q}_{\overline{n}}=\mathrm{softmax}({Q_{\overline{n}}K_{\ast}^{T}}/{\sqrt{d}}){V_{\ast}}$. Finally, we have the masked cross-attention result by simply adding them, that is, $\tilde{Q}_n\oplus \tilde{Q}_{\overline{n}}$. $\oplus$ and $\ominus$ denote element-wise addition and subtraction, respectively. Note that for the UNet to estimate $\epsilon_\ast$, the standard self-attention layer triggered by $s_\ast$ is used instead of the masked cross-attention.\par
The masked cross-attention  accurately puts ``the right objects in the right places;''  the visual information for the $n$-th object is localized around $l_n$ in $\epsilon_n$ and thus the crop-and-merge operation of noises $\{\epsilon_n\}$ guided by $\{l_n\}$ provides $\epsilon$ that accurately reflects the conditions.
Note that the mechanism of NoiseCollage that estimates noise $\epsilon_n$ for each object $n$ independently facilitates the cross-attention between text and image. If we need to process all $N$ objects and their text conditions in the {\em single} cross-attention layer, it is difficult to completely exclude the effects of other $N-1$ objects in the attention process of a certain object. Paint-with-words~\cite{ediff}, a layout-aware text-to-image model based on attention manipulation, tries to control $N$ objects in a single cross-attention layer and often suffers from confusion among the objects. 
%------------------------------------------------------------------
\begin{figure*}[t]
\centering
\includegraphics[width=.9\textwidth]{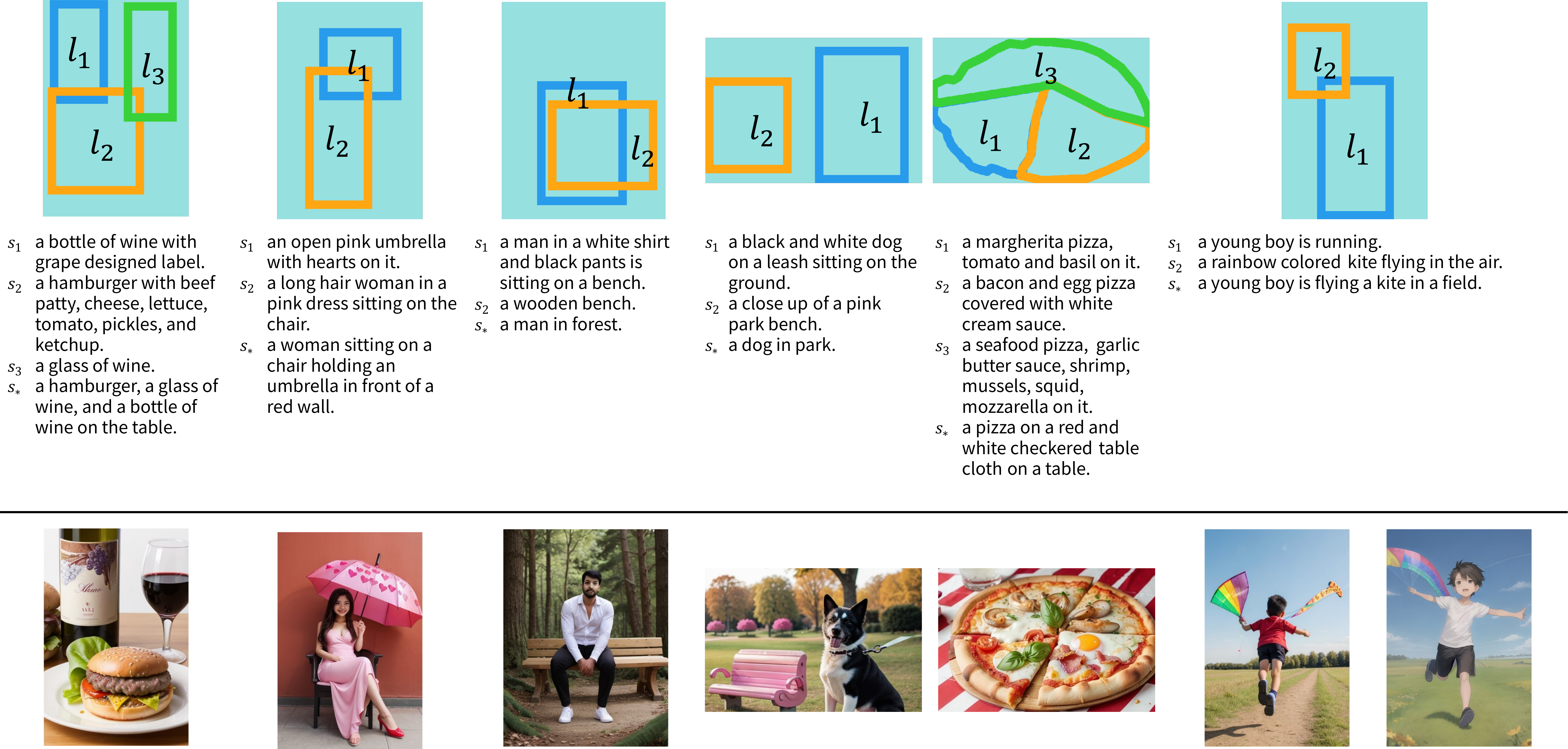}\\[-2mm]
\caption{Images generated by NoiseCollage with layout conditions $L$ and text conditions $(S, s_\ast)$.}
\label{fig:results/example}
\vspace{-2mm}
\end{figure*}

% % MaskedCrossAttentionのテクニック
% \color{blue}
% When the target region is very small compared to the whole region, an image corresponding to the global text condition is sometimes generated in the target region.
% To reduce such failure, we added a global image condition dropout mechanism, named random-dropout.
% It randomly extracts $Q_{uncond}$ from $Q_{bg}$ at a dropout rate (d) and correlates an unconditional text condition to $Q_{uncond}$.
% \color{black}

%==================================================================
\section{Experiments} \label{sec:exp}
%==================================================================
%------------------------------------------------------------------
\subsection{Implementation Details} \label{sec:exp/detail}
We implement NoiseCollage in the SD framework; therefore, the denoising process, including the noise estimation and the crop-and-merge operation, is performed in a latent space; the generated image is given by a decoder from the latent space to the image space. Since NoiseCollage is a training-free model, we employ the pre-trained SD model (SD1.5) by CivitAI\footnote{https://civitai.com/} for generating photo-realistic or anime-style images in the following experiments.
The size of the generated image is set to fit the $512\time 512$ pixel box while keeping its aspect ratio. 
We use UniPCMultistepScheduler~\cite{unipc} as the scheduler of the denoising process and classifier-free guidance~\cite{cfg} with the guidance scale, 7.5. The total denoising step is set to 50. Please refer to the supplementary for the details of the total denoising step and inference step.
% 9:00AM 14th SU
%------------------------------------------------------------------
\subsection{Datasets} \label{sec:exp/dataset}
For performance evaluation experiments, we construct two datasets, BD807 and MD30, where each sample is a combination of an image $x_0$, its layout conditions $L$, and text conditions $(S, s_\ast)$. We collected images from the MS-COCO test dataset because the boundaries of most objects are annotated with polygons and bounding boxes. We use bounding boxes as $L$, which makes NoiseCollage a more handy image generator.
However, later qualitative evaluations use polygons as $L$ in several examples to show the flexibility of the layout condition. We pick up 807 images from the MS-COCO dataset containing $N=2\sim 5$ objects whose region size is larger than $128\times 128$ pixels.\par
Although the MS-COCO dataset also contains image captions, we do not use them as text conditions but prepare our conditions by using BLIP2~\cite{blip2}. This is because the COCO's caption describes the whole image and is inappropriate as the text condition $s_n$ for the individual object. We use the description automatically given by applying BLIP2 to each object region $l_n$ as $s_n$ and the description for the whole image by BLIP2 as $s_\ast$. We call the dataset realized by the above procedure BD807 (BLIP2-guided Dataset with 807 images). MD30 (Manually-annotated Dataset) comprises 30 images chosen from the 807 images. For those images, we discard $s_n$ by BLIP2 and attach a more accurate text as $s_n$ by a human annotator. Note that it only contains 30 images, but its purpose is only to supplement the main result with a larger dataset, BD807.\par
% 19:40 15th Nov. SU

\begin{figure*}[t]
\centering
\includegraphics[width=.95\textwidth]{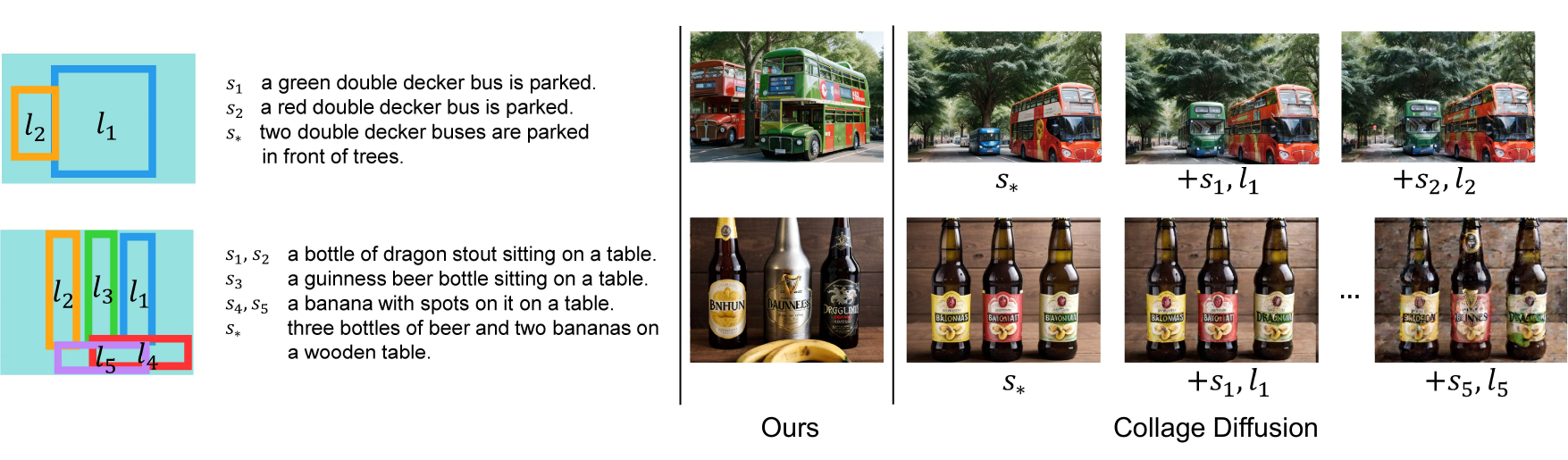}\\[-3mm]
\caption{Comparison of generated images and their generation process by NoiseCollage and Collage Diffusion\cite{collage-diffusion}.}
\label{fig:results/example-comparison-a}
\vspace{-3mm}
\end{figure*}

\begin{figure}[t]
\centering
\includegraphics[width=.95\linewidth]{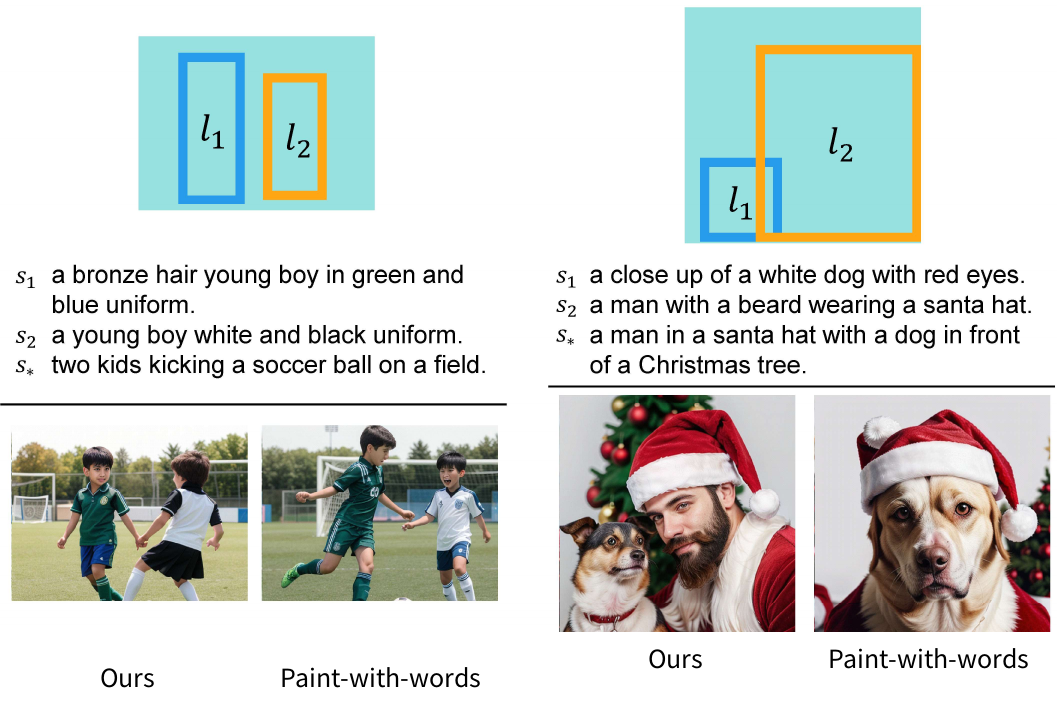}\\[-2mm]
\caption{Comparison of generated images by NoiseCollage and Paint-with-words\cite{ediff}.}
\label{fig:results/example-comparison-b}
\end{figure}

%------------------------------------------------------------------
\subsection{Qualitative Evaluation Result}
\figref{fig:results/example} shows multi-object images generated with various conditions. Layout conditions $L$ are given as bounding boxes or polygons, often overlapping (even largely). Text conditions $S=\{s_1, \ldots, s_N\}$ and $s_{\ast}$ describe the appearance of $N$ objects and the whole image. Note that several conditions in \figref{fig:results/example} are modified from BD807 and MD30 to show the various characteristics of NoiseCollage. \par
The results in \figref{fig:results/example} suggest that the crop-and-merge operation of noises is a very reasonable way to lay out multiple objects accurately. Specifically, it can be said to be ``reasonable'' based on the following two points. First, no artifact exists around the border of the object region $l_n$. Furthermore, a (large) overlap between regions does not degrade the reality of the generated image. 
Second, no confusion exists between layout conditions $\{l_n\}$ and text conditions $\{s_n\}$. In other words, the object described by $s_n$ is correctly located around $l_n$.
Section~\ref{sec:comparison-with-SOTA} shows that even state-of-the-art layout-aware text-to-image models suffer from confusion about the correspondence between texts and locations. 
\par %20:45 14th Nov.
A closer observation of \figref{fig:results/example} reveals various characteristics of NoiseCollage. For example, it shows that we need not be nervous in preparing the global text condition $s_\ast$; for the third and fourth examples from the left, we intentionally use much shorter global text conditions (than the first and second), but the results are still natural. In the pizza image, the layout conditions $L$ are given as polygons; the resulting image shows that polygons help to control the object shapes accurately.
The image of a running boy is generated in two styles, i.e., photo-realistic and anime. NoiseCollage is training-free; thus, any pre-trained noise-estimation model can be plugged into it. This anime-style image is generated simply using a different pre-trained SD model by CivitAI. 
\par % 18:00 16th

%------------------------------------------------------------------
\subsection{Qualitative Comparison with State-of-the-Art Models}
\label{sec:comparison-with-SOTA}
\figref{fig:results/example-comparison-a} compares two generated images and by our NoiseCollage and Collage Diffusion~\cite{collage-diffusion}. Since Collage Diffusion is an iterative editing model, this figure also shows an iterative process where conditions are applied one at a time. Compared to the successful results by NoiseCollage, the results by Collage Diffusion show two issues. The first issue is that the results strongly depend on the initial image given by the global text condition $s_\ast$. In the ``bus'' image, the initial image by $s_\ast$ shows (coincidentally) a red bus on the right side. Then, the second condition $(l_2, s_2)$ is tried to generate a red bus on the left side, but it was ineffective because the red bus is already in the generated image. In the ``bottle'' image, the initial image shows bananas on the label of each bottle. Thus, like the bus image, the fourth and fifth conditions applied later were ineffective.\par
The second issue is the quality degradation by iterations. This degradation becomes more severe when more objects require more iterations. In the ``bottle'' image, the initial image by $s_\ast$ shows readable characters on the bottle labels; however, 
the later iterations gradually degrade their readability, and the characters become almost unreadable in the final image given after five iterations.\par

\figref{fig:results/example-comparison-b} shows comparisons of generated images and their generation process by NoiseCollage and Paint-with-words~\cite{ediff}.
In the ``soccer'' image, Paint-with-words could not correctly color the boys' uniforms, but NoiseCollage succeeded. Paint-with-words assumes shorter text conditions for precise layout control by manipulating its cross-attention module. In other words, such control would be difficult with longer text conditions for describing the detailed appearance of objects. In the ``Santa'' image, two conditions are mixed into one object. This result shows the difficulty of controlling multiple objects in a single cross-attention layer, even with attention manipulation. NoiseCollage uses multiple noises and multiple masked cross-attention operations for individual objects; thus, the objects are well separated.
\begin{table}[t]
\centering
\caption{Average similarity ($\uparrow$) between text conditions $S$ and generated image $x_0$. Red indicates the model with the highest similarity. The parenthesized number ($\uparrow$) shows the percentage of the samples where NoiseCollage shows a better similarity than the comparative model. For example, NoiseCollage outperforms Paint-with-words at 77\% samples of MD30.}\vspace{-2mm} 
% 13:45 14th Nov. SU
\scalebox{.85}[.85]{
\begin{tabular}{l|ccc}
\hline
              & \multicolumn{1}{l}{Paint-with-words} & \multicolumn{1}{l}{Collage Diffusion} & \multicolumn{1}{l}{NoiseCollage}      \\
\hline
                        & 0.250                                & 0.253                                & {\color[HTML]{FE0000} \textbf{0.280}} \\
\multirow{-2}{*}{MD30}  & ( 77\% )                             & ( 70\% )                             &                                \\
                        & 0.240                                & 0.237                                & {\color[HTML]{FE0000} \textbf{0.256}} \\
\multirow{-2}{*}{BD807} & ( 65\% )                             & ( 68\% )                             &                                  \\
\hline
\end{tabular}
}
\label{tab:results/quantitative-comparison}
\end{table}

\begin{figure*}[t]
\centering
\includegraphics[width=.93\linewidth]{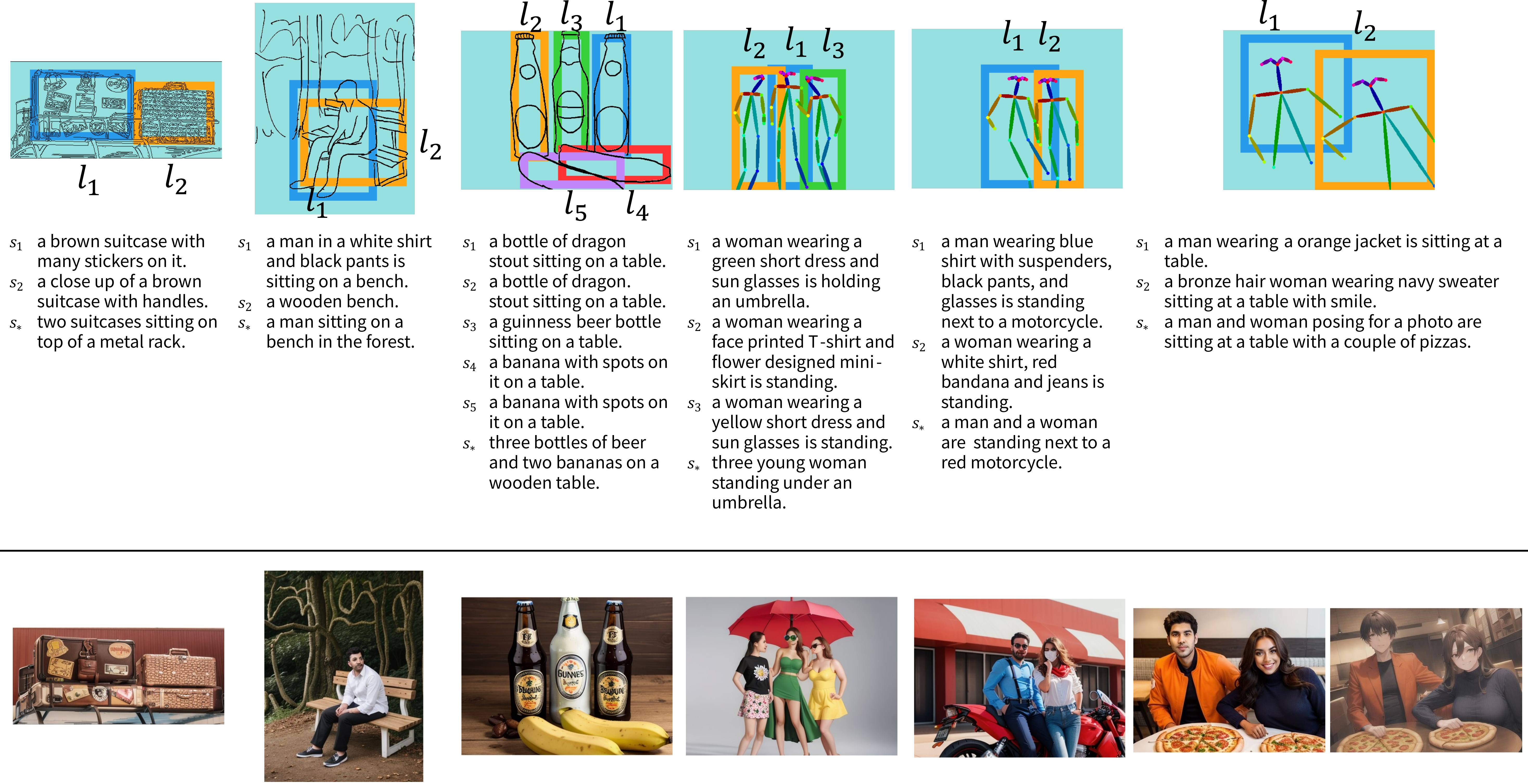}\\[-2mm]
\caption{Images generated by NoiseCollage with ControlNet\cite{controlnet}. The first image is generated with an edge image, the second and third images are with a sketch image, and the remaining images are with a pose skeleton.}
\label{fig:results/example-controlnet}
\end{figure*}

%------------------------------------------------------------------
\subsection{Quantitative Evaluation Results} \label{sec:quantitative}
We evaluate how the generated image accurately reflects the layout and text conditions. Specifically, we evaluate a multimodal similarity between the image region $l_n$ and its text condition $s_n$. If a model appropriately generates an object around $l_n$ while reflecting $s_n$, their multimodal similarity should be high. Following some related works~\cite{collage-diffusion, spatext, Kim2023DenseTG}, we use the ImageEncoder of CLIP~\cite{clip} to have an image feature vector of the region $l_n$ and the TextEncoder for a text feature vector of $s_n$. Then, we use the cosine similarity between these two feature vectors. We have used the same layout and caption conditions described at \ref{sec:exp/dataset} for all the methods for a fair comparison.
% 20:00PM 14th SU
\par
\tabref{tab:results/quantitative-comparison} shows the average similarity achieved by three models (Paint-with-words, CollageDiffution, and NoiseCollage) on the two datasets, MD30 and BD807. In both datasets, NoiseCollage shows higher average similarity than the other models. In the sample-level evaluation, NoiseCollage shows higher similarities than the others in about 70\% samples. These results prove that NoiseCollage can more accurately satisfy the layout and text conditions. Among MD30 and BD807, the latter shows slightly lower similarities; one reason may be that BD807 text conditions are automatically generated.
% 12:25PM 15th Nov. SU

%==================================================================
\section{NoiseCollage with ControlNet}
%==================================================================
%------------------------------------------------------------------
\subsection{Integration of ControlNet for Finer Controls}
ControlNet\cite{controlnet} is a well-known text-to-image diffusion model that can accept various conditions in addition to text conditions. For example, it accepts a pose skeleton to control the pose of a person in the generated image. It also accepts a canny-edge image or a hand-drawn sketch image to control the shape of objects to be generated. The pose skeletons and canny-edge images are automatically generated and the sketch images are created by the authors manually while tracing the images.

\par
% 1:45AM 14th Nov.
%
We can integrate this fine control of ControlNet into NoiseCollage. This integration is done simply by using the pre-trained UNet of ControlNet in the framework of NoiseCollage. The UNet estimates $\epsilon_n$ with an additional condition, such as a pose skeleton, for the $n$-th object. Then, the crop-and-merge operation is performed to have $\epsilon$. Note that $\epsilon_\ast$ is estimated with all the additional conditions and $s_\ast$.
% 7:00AM 15th Nov.
%------------------------------------------------------------------
\subsection{Datasets for Evaluation}
For evaluating the performance of the integrated version, two additional conditions are attached to the dataset of \secref{sec:exp}. Specifically, we prepare a canny-edge image for each image in MD30 and BD807, and a sketch image (drawn manually) for MD30.\par
% 12:45 15th 
%
We prepared two more datasets, HMD20 and HBD256, with human pose as an additional condition. For these datasets, 256 multi-person images are collected from the MS-COCO test dataset. Then, the pose skeleton of each person is estimated by OpenPose~\cite{openpose}. Finally, HBD256 (Human BLIP2 Dataset) is prepared by adding a text condition generated automatically by BLIP2\cite{blip2} for each person region. HMD20 (Human Manual Dataset) is prepared by adding a text condition by a human annotator for each of the 20 images randomly selected from the 256 images.
% 13:00 15th SU
\begin{table}[t]
\centering
\caption{Average similarity ($\uparrow$) in the experiment with ControlNet.  The parenthesized number ($\uparrow$) shows the percentage of the samples where the model shows a better similarity than the other.}
\scalebox{.9}[.9]{
\begin{tabular}{cc|cccc}
\hline
 &  & \multicolumn{2}{c}{Standard} & \multicolumn{2}{c}{NoiseCollage+}   \\
Condition & Dataset & \multicolumn{2}{c}{ControlNet~\cite{controlnet}} & \multicolumn{2}{c}{ControlNet}   \\ \hline
 & RMD30   & 0.293         & (36\%)         & {\color[HTML]{FE0000} \textbf{0.307}} & (64\%)   \\
\multirow{-2}{*}{\begin{tabular}[c]{@{}c@{}} Edge\end{tabular}} & RMD807  & 0.265         & (29\%) & {\color[HTML]{FE0000} \textbf{0.284}} & (71\%)  \\ \hline
Sketch  & RMD30   & 0.276 & (23\%)  & {\color[HTML]{FE0000} \textbf{0.300}}  & (77\%)  \\ \hline
& HMD20   & 0.297         & (15\%)         & {\color[HTML]{FE0000} \textbf{0.332}}    & (85\%)  \\
\multirow{-2}{*}{\begin{tabular}[c]{@{}c@{}}Pose \end{tabular}} & HBD256  & 0.250    & (28\%) & {\color[HTML]{FE0000} \textbf{0.280}}    & (72\%)    \\ \hline
\end{tabular}
}
\label{tab:results/quantitative-comparison-controlnet}
\end{table}

%------------------------------------------------------------------
\subsection{Generated Images by NoiseCollage with ControlNet}
\figref{fig:results/example-controlnet} shows six images generated by NoiseCollage integrated with 
ControlNet. We used ControlNet implemented by Huggingface\footnote{https://huggingface.co/lllyasviel/ControlNet}, and the other details are the same as \secref{sec:exp/detail}.  All those results show multi-object images that accurately reflect the additional conditions to ControlNet. 
For example, the conditions by pose skeletons successfully control the pose of the persons in the generated image. Notably, the integration with ControlNet does not disturb the precise control of NoiseCollage. For example, the third and fourth images from the left accurately reflect their confusing text conditions on bottle types and sunglasses, respectively.

%------------------------------------------------------------------
\subsection{Quantitative Comparison with Standard ControlNet}
\tabref{tab:results/quantitative-comparison-controlnet} shows quantitative evaluation results of the images generated by
NoiseCollage with ControlNet and the standard ControlNet.
The evaluation metric is the multimodal similarity explained in  \secref{sec:quantitative}.
In about 70\% to 80\% of the samples, the performance of ControlNet is improved in our NoiseCollage framework. This improvement is prominent in generating multi-person images with pose skeletons. Although the standard ControlNet can reflect pose conditions in its generated images, it often shows confusion, such as a text condition for one person being reflected in another person. In contrast, if ControlNet is used in NoiseCollage, it can avoid such confusion by estimating noises independently for individual persons under corresponding conditions.
% 16:30  15th SU

%==================================================================
\section{Limitation and Social Impacts}
%==================================================================
\begin{comment}
失敗する例
- オブジェクトが消失する（特にレイアウトが小さい場合）
- オブジェクトが１つになる（単純につながった画像になる場合と特徴が重なる場合）

- テキスト条件の他領域へのリーク
- テキスト条件のドロップアウト
※100%でレイアウトに正確なわけではない（定量評価で関連手法よりこの２つが少ないことは提示している）
\end{comment}

\begin{figure}[t]
\centering
\includegraphics[width=\linewidth]{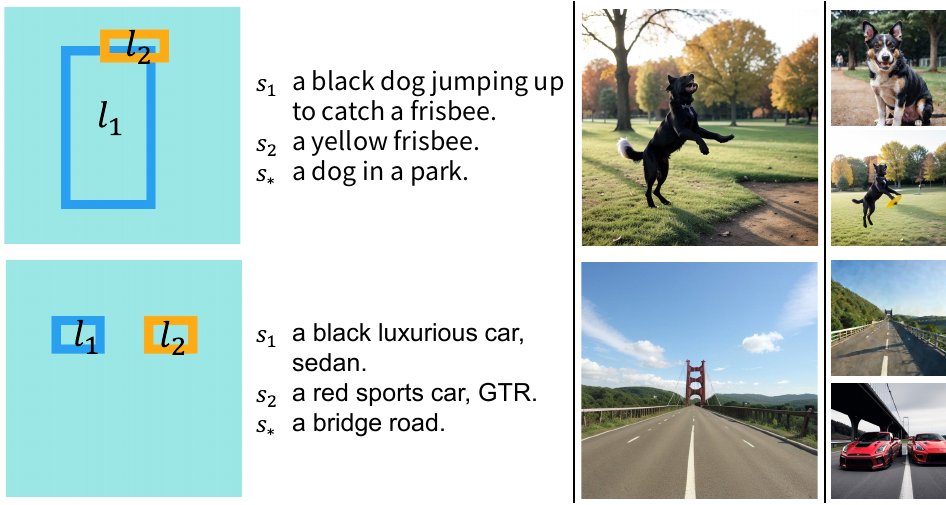}\\[-2mm]
\caption{Failure cases by NoiseCollage. Smaller images on the right side are results by Collage Diffusion~\cite{collage-diffusion} and Paint-with-words~\cite{ediff}.}
\label{fig:results/example-failure}\vspace{-3mm}
\end{figure}

\figref{fig:results/example-failure} shows the limitation of NoiseCollage. NoiseCollage sometimes ignores small objects in the generated image. In the first case, a frisbee is not generated in the image. In the second case, both cars are not generated. Note that the state-of-the-art methods also show difficulty in generating small objects. As shown in the right side of \figref{fig:results/example-failure}, Collage Diffusion also ignores small objects. Paint-with-words did not ignore them but showed them in the wrong places and styles.
\par
Like recent diffusion models, the negative social impact of NoiseCollage is its ability to generate realistic fake images by finer appearance and location control of whole objects or even object parts. For example, since NoiseCollage can independently and easily control individuals in an image, it can potentially create images that depict fake relationships between people.

%==================================================================
\section{Conclusion and Future Work} \label{sec:conclusion}
%==================================================================
This paper proposed a novel layout-aware text-to-image diffusion model called {\em NoiseCollage}. The key idea of NoiseCollage, which can generate multi-object images, is to estimate noises for individual objects independently  and then {\em crop-and-merge} them into a single noise in its denoising process. This operation helps avoid mismatches between the text and layout conditions; in other words, it can accurately put the objects in their right places, while reflecting the text conditions in the corresponding objects.\par 
Qualitative and quantitative evaluations show that NoiseCollage outperforms several state-of-the-art models. These results indicate that the crop-and-merge operation of noises is a reasonable strategy to control image generation. We also show that NoiseCollage can be integrated with ControlNet to use edges, sketches, and pose skeletons as additional conditions. Experimental results show that this integration boosts the layout accuracy of ControlNet.
\par
% 17:50 15th SU
%\color{red}Future work??\color{black}
%
Future work will focus on more efficient layout control. This paper assumes that the layout conditions are given as bounding boxes or polygons. If it is possible to infer possible layout conditions automatically from the given text conditions, it is beneficial for users of NoiseCollage. It is also beneficial if NoiseCollage is extended to accept point annotations, which specify object locations just by points, instead of boxes and polygons. Another research direction is understanding the properties of noise representation against various operations. NoiseCollage shows that cropping and merging (i.e., partial blending) operations realize natural image controls; if we find that applying rigid or non-rigid geometric operations to cropped noises is still possible, we can generate, for example, multi-object videos.  

%==================================================================
\par
\medskip
\noindent{\bf Acknowledgement}\ This work was supported by JSPS (JP22H00540, JP22H05172, JP22H05173).
%==================================================================

\newpage
%% References
{
\small
\bibliographystyle{ieeenat_fullname}
\bibliography{main}
}

\input{X_suppl}

\end{document}

%% file: preamble.tex
\usepackage[dvipsnames]{xcolor}

%% file: X_suppl.tex
\clearpage
\maketitlesupplementary

% \section{Rationale}
% \label{sec:rationale}
% % 
% Having the supplementary compiled together with the main paper means that:
% % 
% \begin{itemize}
% \item The supplementary can back-reference sections of the main paper, for example, we can refer to \cref{sec:intro};
% \item The main paper can forward reference sub-sections within the supplementary explicitly (e.g. referring to a particular experiment); 
% \item When submitted to arXiv, the supplementary will already included at the end of the paper.
% \end{itemize}
% % 
% To split the supplementary pages from the main paper, you can use \href{https://support.apple.com/en-ca/guide/preview/prvw11793/mac#:~:text=Delete%20a%20page%20from%20a,or%20choose%20Edit%20%3E%20Delete).}{Preview (on macOS)}, \href{https://www.adobe.com/acrobat/how-to/delete-pages-from-pdf.html#:~:text=Choose%20%E2%80%9CTools%E2%80%9D%20%3E%20%E2%80%9COrganize,or%20pages%20from%20the%20file.}{Adobe Acrobat} (on all OSs), as well as \href{https://superuser.com/questions/517986/is-it-possible-to-delete-some-pages-of-a-pdf-document}{command line tools}.

This supplementary material shows additional results generated by NoiseCollage. Each of the following figures shows the layout condition $L$ and the text conditions $(S, s_\ast)$, the generated image $x_0$, and 
$N$ individual object images cropped from $x_0$ by $l_1,\ldots, l_N$, from top to bottom. These cropped images not only show the detailed appearance of the individual objects but also show whether the objects are generated at their right place.

% -----------------------------------------------------------------------------
\section{Total Inference step and Time efficiency}
NoiseCollage (also Collage Diffusion~\cite{collage-diffusion}) requires $O(NT)$-times noise estimations, whereas Paint-with-words requires $O(T)$-times where $N$ and $T$ denote the number of objects in a layout and total denoising step, respectively.
Therefore, in NoiseCollage, the total inference step becomes $(N+1)*T$  including noise estimation for the whole image then the number of total inference step increases with the number of objects in a layout.
In fact, NoiseCollage needs 25.7s to generate a single image with $N=5$, whereas Paint-with-words needs 7.42s on a single Nvidia A100GPU.

However, from an optimistic perspective, NoiseCollage can be refined through parallelization, as the $O(N)$-times noise estimations at each time step $t$ are entirely independent. Consequently, it can be executed with $O(T)$ computations.

% -----------------------------------------------------------------------------
\section{Good and bad cases in the results of NoiseCollage}
\begin{figure*}[t]
\centering
\includegraphics[width=\linewidth]{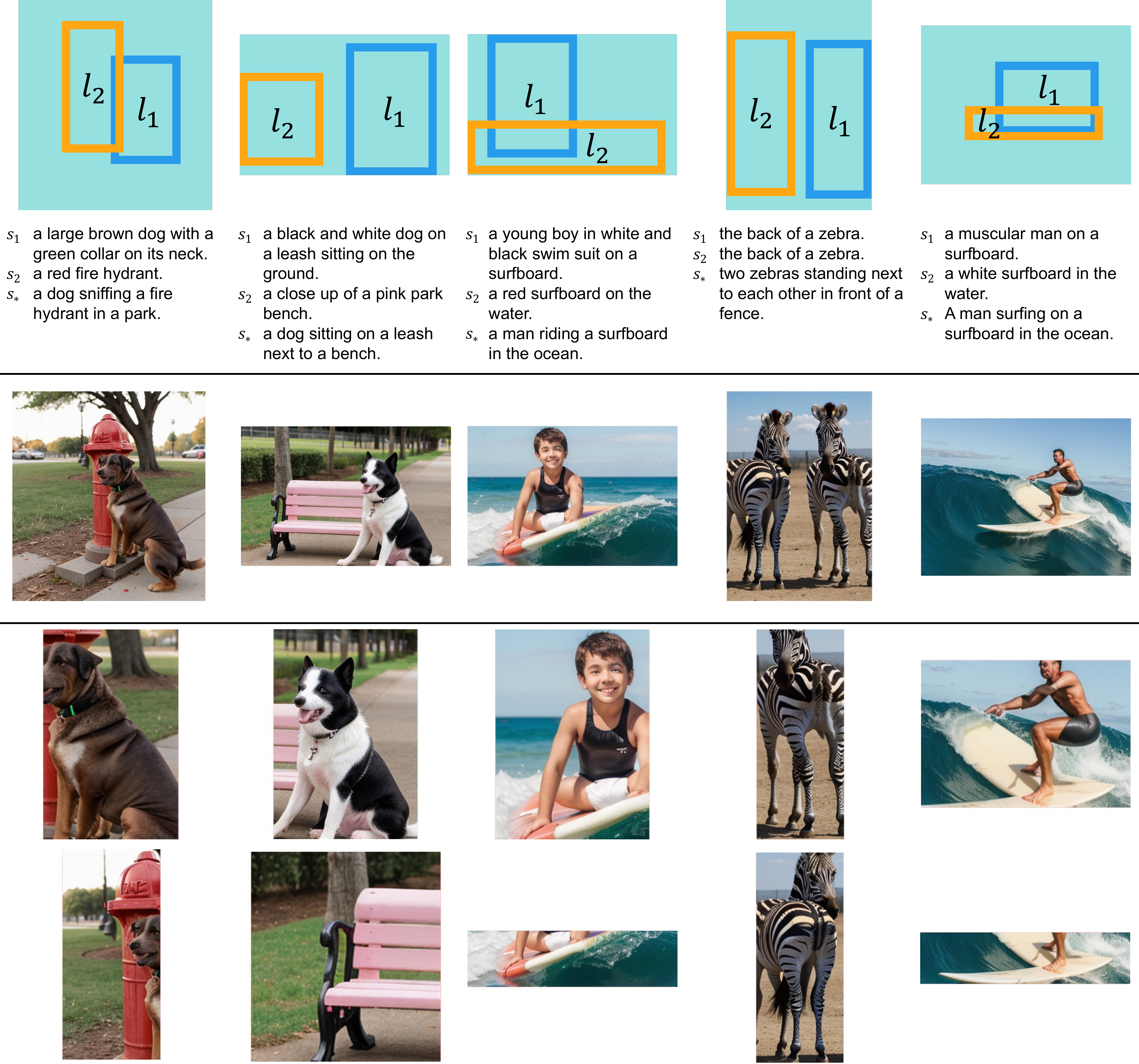}\\[-2mm]
\caption{The best five cases by NoiseCollage on MD30. The lower part shows $N$ individual object images cropped from $x_0$ by $l_1,\ldots, l_N$.}
\label{fig:supp/top5}\vspace{-3mm}
\end{figure*}

\begin{figure*}[t]
\centering
\includegraphics[width=\linewidth]{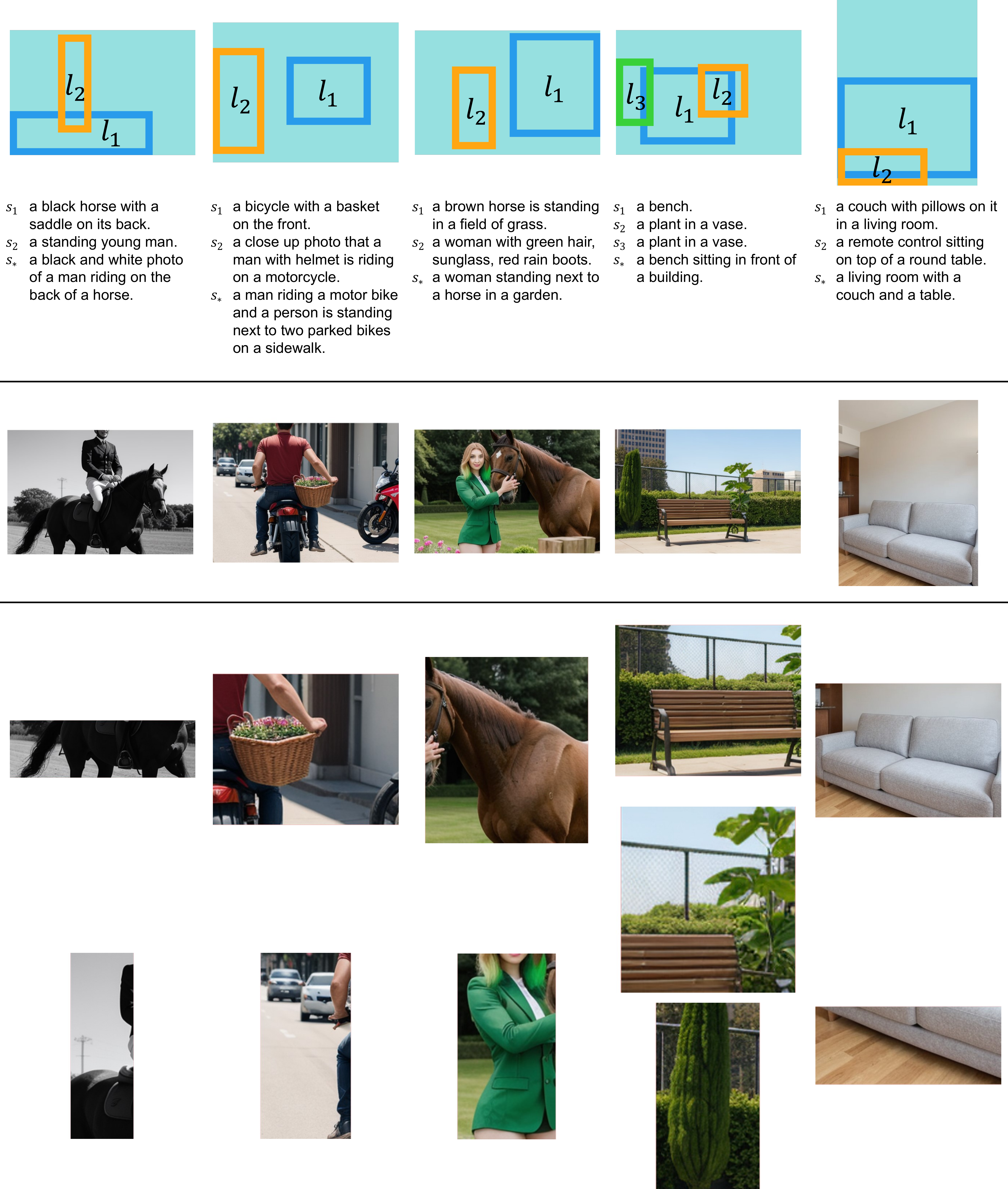}\\[-2mm]
\caption{The worst five cases by NoiseCollage on MD30.}
\label{fig:supp/worst5}\vspace{-3mm}
\end{figure*}

{Figs.~\color{red}\ref{fig:supp/top5}} and {\color{red}\ref{fig:supp/worst5}} show the images generated by NoiseCollage on the MD30 dataset. The former shows images with good scores, while the latter shows images with bad scores, according to the evaluation metric (multimodal similarity between the $n$-th object image and its text condition $s_n$) of \secref{sec:quantitative}. The good cases of \figref{fig:supp/top5} show the accurate correspondence between $s_n$ and $l_n$, even the layout $l_n$ is specified by a bounding box.
\par
Even in the worst cases of \figref{fig:supp/worst5}, large objects (specified by $l_1$ and $s_1$) are generated at the right place with a correct appearance (except for the bicycle image). Except for the remote control that disappears from the generated image, small objects are generated in misaligned locations and still look good.

% -----------------------------------------------------------------------------
\section{More Results of NoiseCollage with ControlNet}
\begin{figure*}[t]
\centering
\includegraphics[width=\linewidth]{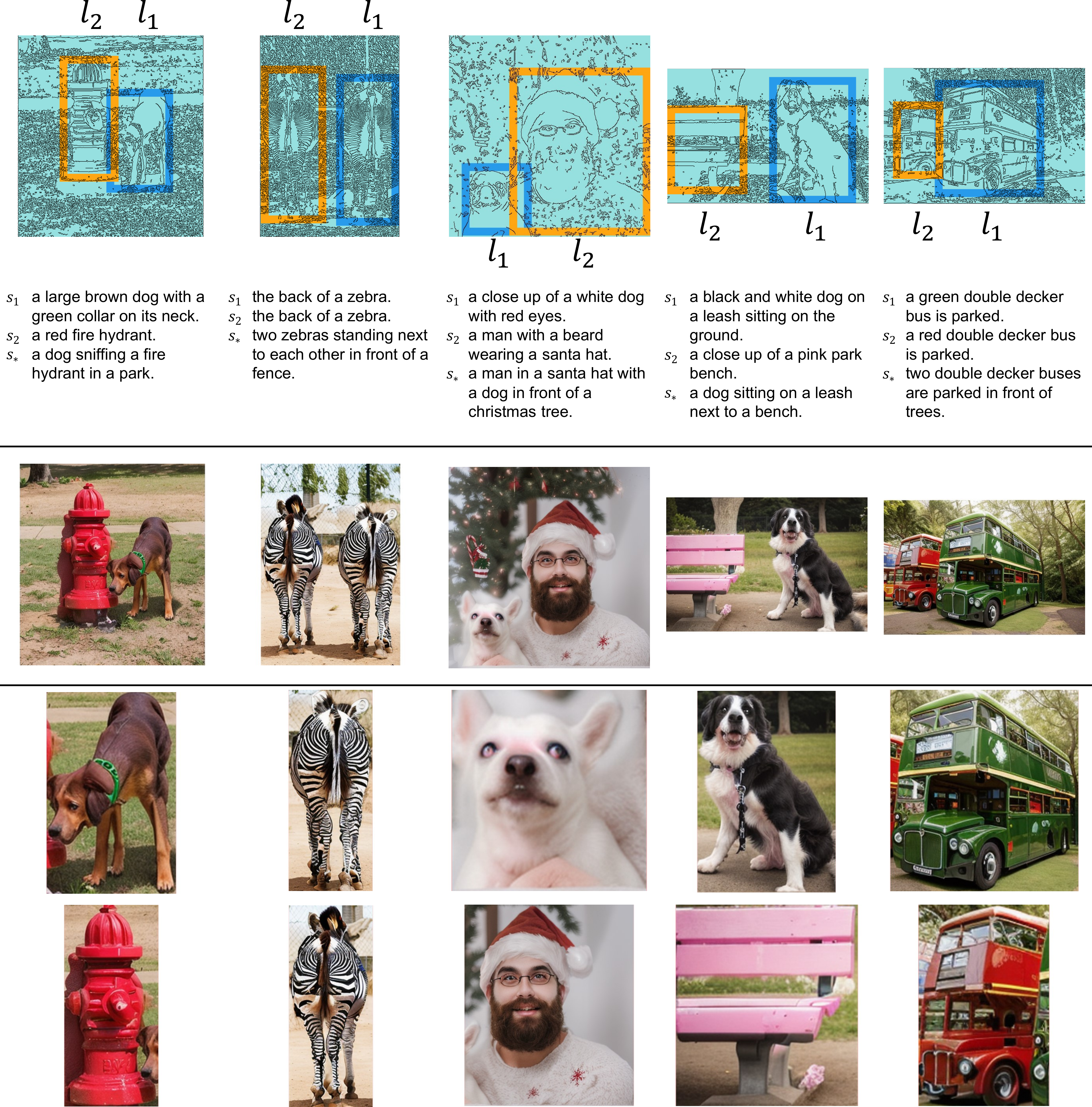}\\[-2mm]
\caption{Images generated by NoiseCollage with ControlNet\cite{controlnet} of an edge image condition.}
\label{fig:supp/controlnet-canny}\vspace{-3mm}
\end{figure*}

\begin{figure*}[t]
\centering
\includegraphics[width=\linewidth]{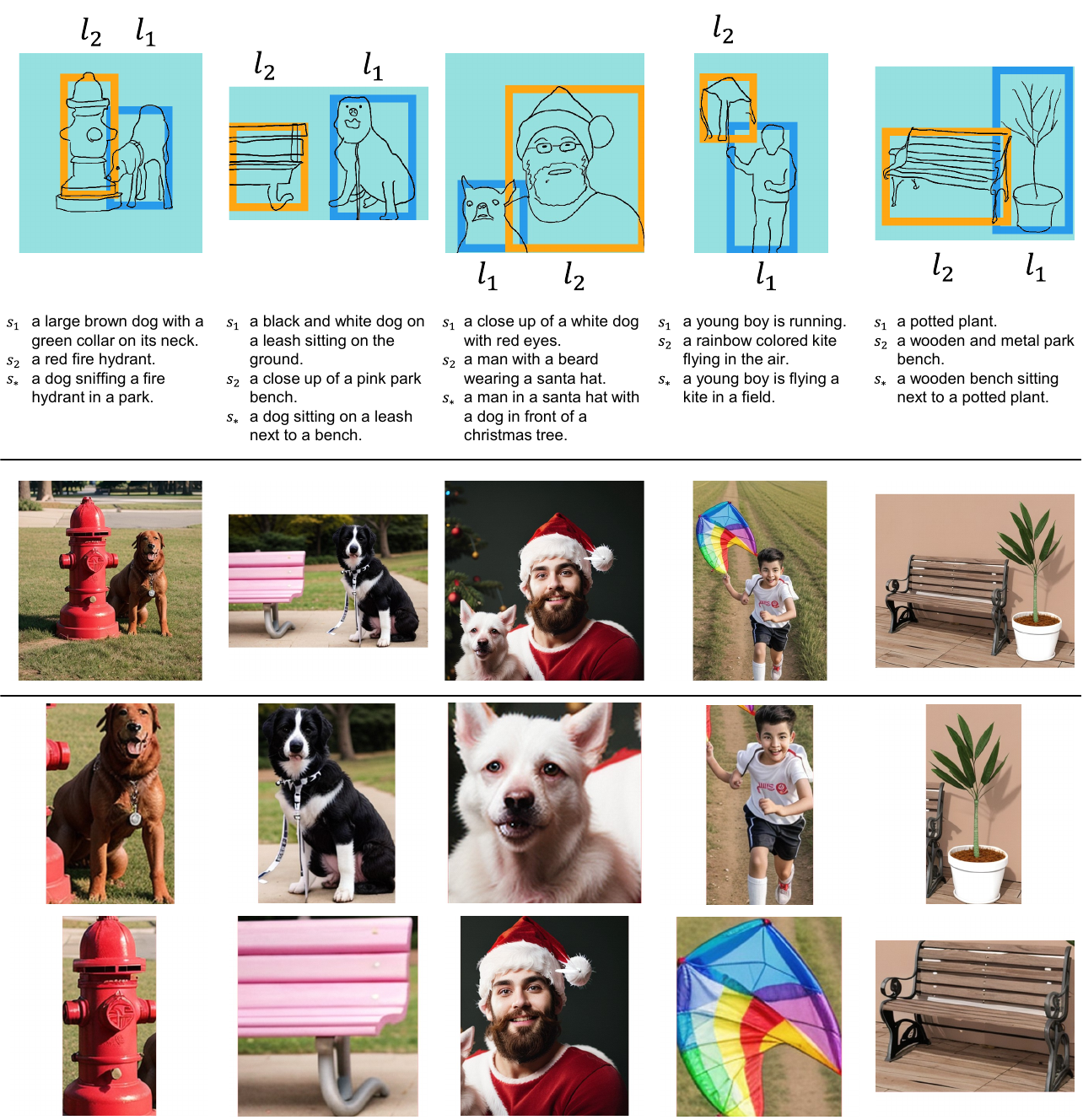}\\[-2mm]
\caption{Images generated by NoiseCollage with ControlNet\cite{controlnet} of a sketch condition.}
\label{fig:supp/controlnet-sketch}\vspace{-3mm}
\end{figure*}

\begin{figure*}[t]
\centering
\includegraphics[width=.9\linewidth]{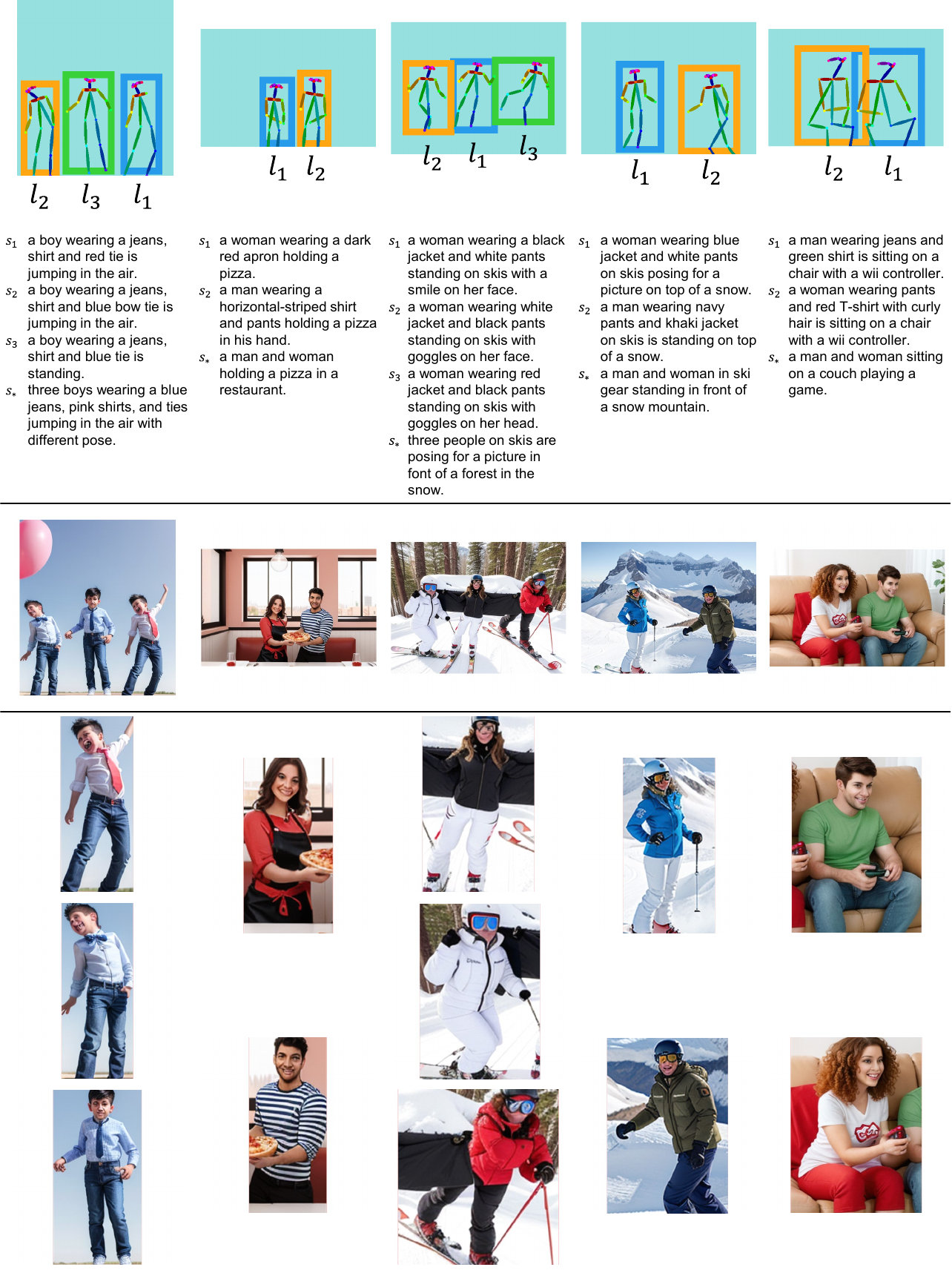}\\[-2mm]
\caption{Images generated by NoiseCollage with ControlNet\cite{controlnet} of a pose skeleton condition.}
\label{fig:supp/controlnet-pose}\vspace{-3mm}
\end{figure*}

While we already showed several results of NoiseCollage with ControlNet~\cite{controlnet} in \figref{fig:results/example-controlnet}, we show more results in {Figs.~\color{red}\ref{fig:supp/controlnet-canny}}, {\color{red}\ref{fig:supp/controlnet-sketch}}, and
{\color{red}\ref{fig:supp/controlnet-pose}}, which uses edge images, sketches, and pose skeletons as additional constraints, respectively. Like the results in \figref{fig:results/example-controlnet}, the additional results also show how the conditions for ControlNet guide the output images accurately.
Note that bounding boxes specify the layout conditions, whereas polygons are used in \figref{fig:results/example-controlnet}. We can confirm that bounding boxes are also easy but appropriate layout conditions.

% -----------------------------------------------------------------------------
\section{Results of more crowded layouts}
As already stated in the ``Limitations'' section, it is difficult for not only ours but also baselines to generate images under complex layouts with small objects or a large number of objects. This limitation may come from the fact that the common backbone, StableDiffusion~\cite{sd}, uses a low-resolution latent space of $64\times 64$.

\begin{figure*}[t]
\centering
\includegraphics[width=\linewidth]{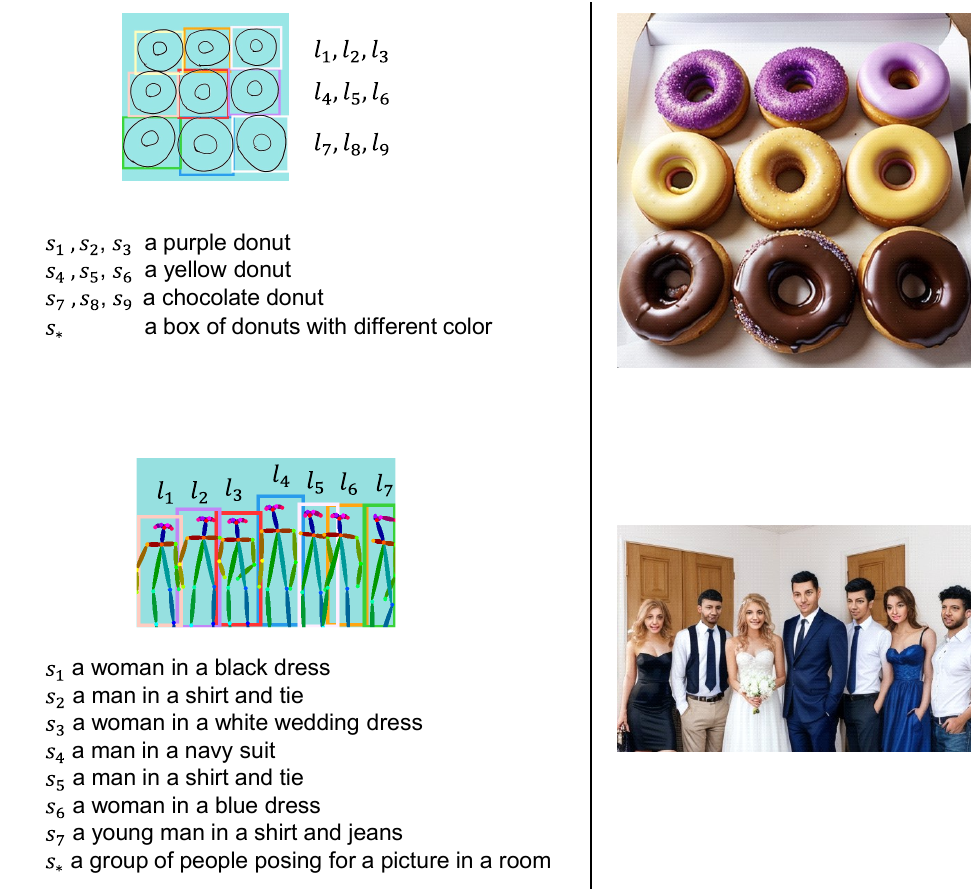}\\[-2mm]
\caption{Generated images with more complex layouts. Ours can control the layout of donuts and the clothes of each person.}
\label{fig:reb/complex}
\vspace{-3mm}
\end{figure*}

\figref{fig:reb/complex} shows the generated images with more crowded objects ($N=7,9$) by NoiseCollage with ControlNet. In these examples, the layout conditions are well reflected in the resulting images. However, if we want to put more objects, say, $N=20$, it is difficult to expect the accurate reflection of their conditions, as noted above.

\clearpage

% Top 5.

% Worst 5.

% % comparison vs CD
% \begin{figure*}[t]
% \centering
% \includegraphics[width=.7\linewidth]{_supp_/comp-cd.pdf}\\[-2mm]
% \caption{Comparison of generated images by Collage Diffusion\cite{collage-diffusion} and NoiseCollage.}
% \label{fig:supp/comp-cd}\vspace{-3mm}
% \end{figure*}

% % comparison vs Paint
% \begin{figure*}[t]
% \centering
% \includegraphics[width=.7\linewidth]{_supp_/comp-paint.pdf}\\[-2mm]
% \caption{Comparison of generated images by Paint-with-words\cite{ediff} and NoiseCollage.}
% \label{fig:supp/comp-paint}\vspace{-3mm}
% \end{figure*}

% ControlNet - canny

% ControlNet - sketch

% ControlNet - pose

% ControlNet - pose (comparison)
% \begin{figure*}[t]
% \centering
% \includegraphics[width=\linewidth]{_supp_/controlnet-pose-detail.pdf}\\[-2mm]
% \caption{Comparison of generated images by ControlNet\cite{controlnet} of a pose skeleton condition with and without NoiseCollage.}
% \label{fig:supp/controlnet-pose-detail}\vspace{-3mm}
% \end{figure*}

% Revised version (入れれないかも？)
% \begin{figure*}[t]
% \centering
% \includegraphics[width=\linewidth]{_supp_/example-controlnet-fixed.pdf}\\[-2mm]
% \caption{The revised version of \figref{fig:results/example-controlnet}. these images are generated from the layout input conditions given as bounding boxes, not polygons.}
% \label{fig:supp/controlnet-pose}\vspace{-3mm}
% \end{figure*}

%% file: main.bbl
\begin{thebibliography}{43}
\providecommand{\natexlab}[1]{#1}
\providecommand{\url}[1]{\texttt{#1}}
\expandafter\ifx\csname urlstyle\endcsname\relax
  \providecommand{\doi}[1]{doi: #1}\else
  \providecommand{\doi}{doi: \begingroup \urlstyle{rm}\Url}\fi

\bibitem[Avrahami et~al.(2021)Avrahami, Lischinski, and Fried]{blend-dm}
Omri Avrahami, Dani Lischinski, and Ohad Fried.
\newblock {Blended Diffusion for Text-driven Editing of Natural Images}.
\newblock \emph{2022 IEEE/CVF Conference on Computer Vision and Pattern Recognition (CVPR)}, pages 18187--18197, 2021.

\bibitem[Avrahami et~al.(2022)Avrahami, Fried, and Lischinski]{blend-ldm}
Omri Avrahami, Ohad Fried, and Dani Lischinski.
\newblock {Blended latent diffusion}.
\newblock \emph{ACM Transactions on Graphics (TOG)}, 42\penalty0 (4):\penalty0 1--11, 2022.

\bibitem[Avrahami et~al.(2023)Avrahami, Hayes, Gafni, Gupta, Taigman, Parikh, Lischinski, Fried, and Yin]{spatext}
Omri Avrahami, Thomas Hayes, Oran Gafni, Sonal Gupta, Yaniv Taigman, Devi Parikh, Dani Lischinski, Ohad Fried, and Xiaoyue Yin.
\newblock {Spatext: Spatio-textual representation for controllable image generation}.
\newblock In \emph{Proceedings of the IEEE/CVF Conference on Computer Vision and Pattern Recognition}, pages 18370--18380, 2023.

\bibitem[Balaji et~al.(2022)Balaji, Nah, Huang, Vahdat, Song, Zhang, Kreis, Aittala, Aila, Laine, Catanzaro, Karras, and Liu]{ediff}
Yogesh Balaji, Seungjun Nah, Xun Huang, Arash Vahdat, Jiaming Song, Qinsheng Zhang, Karsten Kreis, Miika Aittala, Timo Aila, Samuli Laine, Bryan Catanzaro, Tero Karras, and Ming-Yu Liu.
\newblock {ediffi: Text-to-image diffusion models with an ensemble of expert denoisers}.
\newblock \emph{arXiv preprint arXiv:2211.01324}, 2022.

\bibitem[Brooks et~al.(2023)Brooks, Holynski, and Efros]{Brooks2022InstructPix2PixLT}
Tim Brooks, Aleksander Holynski, and Alexei~A. Efros.
\newblock {Instructpix2pix: Learning to follow image editing instructions}.
\newblock In \emph{Proceedings of the IEEE/CVF Conference on Computer Vision and Pattern Recognition}, pages 18392--18402, 2023.

\bibitem[Cao et~al.(2018)Cao, Hidalgo, Simon, Wei, and Sheikh]{openpose}
Zhe Cao, Gines Hidalgo, Tomas Simon, Shih-En Wei, and Yaser Sheikh.
\newblock {OpenPose: Realtime Multi-Person 2D Pose Estimation Using Part Affinity Fields}.
\newblock \emph{IEEE Transactions on Pattern Analysis and Machine Intelligence}, 43:\penalty0 172--186, 2018.

\bibitem[Cheng et~al.(2023)Cheng, Liang, Shi, He, Xiao, and Li]{Cheng2023LayoutDiffuseAF}
Jiaxin Cheng, Xiao Liang, Xingjian Shi, Tong He, Tianjun Xiao, and Mu Li.
\newblock {Layoutdiffuse: Adapting foundational diffusion models for layout-to-image generation}.
\newblock \emph{arXiv preprint arXiv:2302.08908}, 2023.

\bibitem[Couairon et~al.(2022)Couairon, Verbeek, Schwenk, and Cord]{Couairon2022DiffEditDS}
Guillaume Couairon, Jakob Verbeek, Holger Schwenk, and Matthieu Cord.
\newblock {Diffedit: Diffusion-based semantic image editing with mask guidance}.
\newblock \emph{arXiv preprint arXiv:2210.11427}, 2022.

\bibitem[Couairon et~al.(2023)Couairon, Careil, Cord, Lathuili{\`e}re, and Verbeek]{zestguide}
Guillaume Couairon, Marlene Careil, Matthieu Cord, St{\'e}phane Lathuili{\`e}re, and Jakob Verbeek.
\newblock {Zero-shot spatial layout conditioning for text-to-image diffusion models}.
\newblock In \emph{Proceedings of the IEEE/CVF International Conference on Computer Vision}, pages 2174--2183, 2023.

\bibitem[Dhariwal and Nichol(2021)]{dmbg}
Prafulla Dhariwal and Alex Nichol.
\newblock {Diffusion Models Beat GANs on Image Synthesis}.
\newblock \emph{arXiv preprint arxiv:2105.05233}, 2021.

\bibitem[Gafni et~al.(2022)Gafni, Polyak, Ashual, Sheynin, Parikh, and Taigman]{make-a-scene}
Oran Gafni, Adam Polyak, Oron Ashual, Shelly Sheynin, Devi Parikh, and Yaniv Taigman.
\newblock {Make-a-scene: Scene-based text-to-image generation with human priors}.
\newblock In \emph{European Conference on Computer Vision}, pages 89--106, 2022.

\bibitem[Hertz et~al.(2022)Hertz, Mokady, Tenenbaum, Aberman, Pritch, and Cohen-Or]{Hertz2022PrompttoPromptIE}
Amir Hertz, Ron Mokady, Jay~M. Tenenbaum, Kfir Aberman, Yael Pritch, and Daniel Cohen-Or.
\newblock {Prompt-to-prompt image editing with cross attention control}.
\newblock \emph{arXiv preprint arXiv:2208.01626}, 2022.

\bibitem[Ho and Salimans(2022)]{cfg}
Jonathan Ho and Tim Salimans.
\newblock {Classifier-free diffusion guidance}.
\newblock \emph{arXiv preprint arXiv:2207.12598}, 2022.

\bibitem[Ho et~al.(2020)Ho, Jain, and Abbeel]{ddpm}
Jonathan Ho, Ajay Jain, and P. Abbeel.
\newblock {Denoising diffusion probabilistic models}.
\newblock \emph{Advances in Neural Information Processing Systems}, 33:\penalty0 6840--6851, 2020.

\bibitem[Jia et~al.(2023)Jia, Luo, Dang, Dai, Chang, Wang, and Wang]{Jia2023SSMGSM}
Chengyou Jia, Minnan Luo, Zhuohang Dang, Guangwen Dai, Xiaojun Chang, Mengmeng Wang, and Jingdong Wang.
\newblock {SSMG: Spatial-Semantic Map Guided Diffusion Model for Free-form Layout-to-Image Generation}.
\newblock \emph{arXiv preprint arXiv:2308.10156}, 2023.

\bibitem[Kim et~al.(2023)Kim, Lee, Kim, Ha, and Zhu]{Kim2023DenseTG}
Yunji Kim, Jiyoung Lee, Jin-Hwa Kim, Jung-Woo Ha, and Jun-Yan Zhu.
\newblock {Dense text-to-image generation with attention modulation}.
\newblock In \emph{Proceedings of the IEEE/CVF International Conference on Computer Vision}, pages 7701--7711, 2023.

\bibitem[Li et~al.(2023)Li, Li, Savarese, and Hoi]{blip2}
Junnan Li, Dongxu Li, Silvio Savarese, and Steven C.~H. Hoi.
\newblock {Blip-2: Bootstrapping language-image pre-training with frozen image encoders and large language models}.
\newblock \emph{arXiv preprint arXiv:2301.12597}, 2023.

\bibitem[Lugmayr et~al.(2022)Lugmayr, Danelljan, Romero, Yu, Timofte, and Gool]{Lugmayr2022RePaintIU}
Andreas Lugmayr, Martin Danelljan, Andr{\'e}s Romero, Fisher Yu, Radu Timofte, and Luc~Van Gool.
\newblock {Repaint: Inpainting using denoising diffusion probabilistic models}.
\newblock In \emph{Proceedings of the IEEE/CVF Conference on Computer Vision and Pattern Recognition}, pages 11461--11471, 2022.

\bibitem[Mao and Wang(2023)]{layout-cyber1}
Jiafeng Mao and Xueting Wang.
\newblock {Training-Free Location-Aware Text-to-Image Synthesis}.
\newblock \emph{arXiv preprint arXiv:2304.13427}, 2023.

\bibitem[Mao et~al.(2023)Mao, Wang, and Aizawa]{layout-cyber2}
Jiafeng Mao, Xueting Wang, and Kiyoharu Aizawa.
\newblock {Guided Image Synthesis via Initial Image Editing in Diffusion Model}.
\newblock \emph{Proceedings of the 31st ACM International Conference on Multimedia}, 2023.

\bibitem[Meng et~al.(2021)Meng, He, Song, Song, Wu, Zhu, and Ermon]{sdedit}
Chenlin Meng, Yutong He, Yang Song, Jiaming Song, Jiajun Wu, Jun-Yan Zhu, and Stefano Ermon.
\newblock {SDEdit: Guided Image Synthesis and Editing with Stochastic Differential Equations}.
\newblock In \emph{International Conference on Learning Representations}, 2021.

\bibitem[Nichol and Dhariwal(2021)]{improved-ddpm}
Alex Nichol and Prafulla Dhariwal.
\newblock {Improved denoising diffusion probabilistic models}.
\newblock In \emph{International Conference on Machine Learning}, pages 8162--8171, 2021.

\bibitem[Parmar et~al.(2023)Parmar, Singh, Zhang, Li, Lu, and Zhu]{Parmar2023ZeroshotIT}
Gaurav Parmar, Krishna~Kumar Singh, Richard Zhang, Yijun Li, Jingwan Lu, and Jun-Yan Zhu.
\newblock {Zero-shot image-to-image translation}.
\newblock In \emph{ACM SIGGRAPH 2023 Conference Proceedings}, pages 1--11, 2023.

\bibitem[Phung et~al.(2023)Phung, Ge, and Huang]{Phung2023GroundedTS}
Quynh Phung, Songwei Ge, and Jia-Bin Huang.
\newblock {Grounded Text-to-Image Synthesis with Attention Refocusing}.
\newblock \emph{arXiv preprint arXiv:2306.05427}, 2023.

\bibitem[Podell et~al.(2023)Podell, English, Lacey, Blattmann, Dockhorn, Muller, Penna, and Rombach]{sdxl}
Dustin Podell, Zion English, Kyle Lacey, A. Blattmann, Tim Dockhorn, Jonas Muller, Joe Penna, and Robin Rombach.
\newblock {Sdxl: Improving latent diffusion models for high-resolution image synthesis}.
\newblock \emph{arXiv preprint arXiv:2307.01952}, 2023.

\bibitem[Radford et~al.(2021)Radford, Kim, Hallacy, Ramesh, Goh, Agarwal, Sastry, Askell, Mishkin, Clark, Krueger, and Sutskever]{clip}
Alec Radford, Jong~Wook Kim, Chris Hallacy, Aditya Ramesh, Gabriel Goh, Sandhini Agarwal, Girish Sastry, Amanda Askell, Pamela Mishkin, Jack Clark, Gretchen Krueger, and Ilya Sutskever.
\newblock {Learning Transferable Visual Models From Natural Language Supervision}.
\newblock In \emph{International Conference on Machine Learning}, 2021.

\bibitem[Ramesh et~al.(2022)Ramesh, Dhariwal, Nichol, Chu, and Chen]{dalle2}
Aditya Ramesh, Prafulla Dhariwal, Alex Nichol, Casey Chu, and Mark Chen.
\newblock {Hierarchical text-conditional image generation with clip latents}.
\newblock \emph{arXiv preprint arXiv:2204.06125}, 1\penalty0 (2):\penalty0 3, 2022.

\bibitem[Rombach et~al.(2022)Rombach, Blattmann, Lorenz, Esser, and Ommer]{sd}
Robin Rombach, A. Blattmann, Dominik Lorenz, Patrick Esser, and Bj{\"o}rn Ommer.
\newblock {High-resolution image synthesis with latent diffusion models}.
\newblock In \emph{Proceedings of the IEEE/CVF Conference on Computer Vision and Pattern Recognition}, pages 10684--10695, 2022.

\bibitem[Saharia et~al.(2022{\natexlab{a}})Saharia, Chan, Chang, Lee, Ho, Salimans, Fleet, and Norouzi]{palette}
Chitwan Saharia, William Chan, Huiwen Chang, Chris~A. Lee, Jonathan Ho, Tim Salimans, David~J. Fleet, and Mohammad Norouzi.
\newblock {Palette: Image-to-image diffusion models}.
\newblock In \emph{ACM SIGGRAPH 2022 Conference Proceedings}, pages 1--10, 2022{\natexlab{a}}.

\bibitem[Saharia et~al.(2022{\natexlab{b}})Saharia, Chan, Saxena, Li, Whang, Denton, Ghasemipour, Ayan, Mahdavi, Lopes, Salimans, Ho, Fleet, and Norouzi]{imagen}
Chitwan Saharia, William Chan, Saurabh Saxena, Lala Li, Jay Whang, Emily~L. Denton, Seyed Kamyar~Seyed Ghasemipour, Burcu~Karagol Ayan, Seyedeh~Sara Mahdavi, Raphael~Gontijo Lopes, Tim Salimans, Jonathan Ho, David~J. Fleet, and Mohammad Norouzi.
\newblock {Photorealistic text-to-image diffusion models with deep language understanding}.
\newblock \emph{Advances in Neural Information Processing Systems}, 35:\penalty0 36479--36494, 2022{\natexlab{b}}.

\bibitem[Sarukkai et~al.(2023)Sarukkai, Li, Ma, R'e, and Fatahalian]{collage-diffusion}
Vishnu Sarukkai, Linden Li, Arden Ma, Christopher R'e, and Kayvon Fatahalian.
\newblock {Collage diffusion}.
\newblock \emph{arXiv preprint arXiv:2303.00262}, 2023.

\bibitem[Sohl-Dickstein et~al.(2015)Sohl-Dickstein, Weiss, Maheswaranathan, and Ganguli]{ddpm-org}
Jascha~Narain Sohl-Dickstein, Eric~A. Weiss, Niru Maheswaranathan, and Surya Ganguli.
\newblock {Deep unsupervised learning using nonequilibrium thermodynamics}.
\newblock In \emph{International Conference on Machine Learning}, pages 2256--2265, 2015.

\bibitem[Song et~al.(2020)Song, Meng, and Ermon]{ddim}
Jiaming Song, Chenlin Meng, and Stefano Ermon.
\newblock {Denoising diffusion implicit models}.
\newblock \emph{arXiv preprint arXiv:2010.02502}, 2020.

\bibitem[Xiao et~al.(2023)Xiao, Li, Lv, Wang, and Huang]{Xiao2023RBRA}
Jiayu Xiao, Liang Li, Henglei Lv, Shuhui Wang, and Qingming Huang.
\newblock {R\&B: Region and Boundary Aware Zero-shot Grounded Text-to-image Generation}.
\newblock \emph{arXiv preprint arXiv:2310.08872}, 2023.

\bibitem[Xie et~al.(2023{\natexlab{a}})Xie, Li, Huang, Liu, Zhang, Zheng, and Shou]{boxdiffusion}
Jinheng Xie, Yuexiang Li, Yawen Huang, Haozhe Liu, Wentian Zhang, Yefeng Zheng, and Mike~Zheng Shou.
\newblock {BoxDiff: Text-to-Image Synthesis with Training-Free Box-Constrained Diffusion}.
\newblock In \emph{Proceedings of the IEEE/CVF International Conference on Computer Vision}, pages 7452--7461, 2023{\natexlab{a}}.

\bibitem[Xie et~al.(2023{\natexlab{b}})Xie, Zhang, Lin, Hinz, and Zhang]{Xie2022SmartBrushTA}
Shaoan Xie, Zhifei Zhang, Zhe Lin, Tobias Hinz, and Kun Zhang.
\newblock {Smartbrush: Text and shape guided object inpainting with diffusion model}.
\newblock In \emph{Proceedings of the IEEE/CVF Conference on Computer Vision and Pattern Recognition}, pages 22428--22437, 2023{\natexlab{b}}.

\bibitem[Xue et~al.(2023)Xue, Huang, Sun, Song, and Zhang]{Xue2023FreestyleLS}
Han Xue, Zhi~Feng Huang, Qianru Sun, Li Song, and Wenjun Zhang.
\newblock {Freestyle Layout-to-Image Synthesis}.
\newblock In \emph{Proceedings of the IEEE/CVF Conference on Computer Vision and Pattern Recognition}, pages 14256--14266, 2023.

\bibitem[Yang et~al.(2023)Yang, Gu, Zhang, Zhang, Chen, Sun, Chen, and Wen]{paint-by-example}
Binxin Yang, Shuyang Gu, Bo Zhang, Ting Zhang, Xuejin Chen, Xiaoyan Sun, Dong Chen, and Fang Wen.
\newblock {Paint by example: Exemplar-based image editing with diffusion models}.
\newblock In \emph{Proceedings of the IEEE/CVF Conference on Computer Vision and Pattern Recognition}, pages 18381--18391, 2023.

\bibitem[Zhang et~al.(2023{\natexlab{a}})Zhang, Ji, Zhang, Yu, Jaakkola, and Chang]{Zhang2023TowardsCI}
Guanhua Zhang, Jiabao Ji, Yang Zhang, Mo Yu, T. Jaakkola, and Shiyu Chang.
\newblock {Towards Coherent Image Inpainting Using Denoising Diffusion Implicit Models}.
\newblock In \emph{International Conference on Machine Learning}, 2023{\natexlab{a}}.

\bibitem[Zhang et~al.(2023{\natexlab{b}})Zhang, Rao, and Agrawala]{controlnet}
Lvmin Zhang, Anyi Rao, and Maneesh Agrawala.
\newblock {Adding conditional control to text-to-image diffusion models}.
\newblock In \emph{Proceedings of the IEEE/CVF International Conference on Computer Vision}, pages 3836--3847, 2023{\natexlab{b}}.

\bibitem[Zhang et~al.(2023{\natexlab{c}})Zhang, Guo, Yoo, Matsuo, and Iwasawa]{Zhang2023paste}
X. Zhang, Jiaxian Guo, Paul Yoo, Yutaka Matsuo, and Yusuke Iwasawa.
\newblock {Paste, Inpaint and Harmonize via Denoising: Subject-Driven Image Editing with Pre-Trained Diffusion Model}.
\newblock \emph{arXiv preprint arXiv:2306.07596}, 2023{\natexlab{c}}.

\bibitem[Zhang et~al.(2023{\natexlab{d}})Zhang, Huang, and Liao]{continuous-layout}
Zhiyuan Zhang, Zhitong Huang, and Jingtang Liao.
\newblock {Continuous Layout Editing of Single Images with Diffusion Models}.
\newblock \emph{arXiv preprint arXiv:2306.13078}, 2023{\natexlab{d}}.

\bibitem[Zhao et~al.(2023)Zhao, Bai, Rao, Zhou, and Lu]{unipc}
Wenliang Zhao, Lujia Bai, Yongming Rao, Jie Zhou, and Jiwen Lu.
\newblock {UniPC: A Unified Predictor-Corrector Framework for Fast Sampling of Diffusion Models}.
\newblock \emph{arXiv preprint arXiv:2302.04867}, 2023.

\end{thebibliography}
